\documentclass[notoc, tikz]{tufte-handout}

\usepackage{amssymb,amsmath,amsthm}
\usepackage{bbm}
\usepackage{xifthen}
\usepackage{natbib}
\usepackage{xcolor}
\usepackage{colortbl}
\usepackage{mdframed}
\usepackage{graphicx} 
\usepackage{accents}
\usepackage{booktabs}
\usepackage{nicefrac}
\usepackage{tabu} %
\usepackage{caption}
\usepackage{cite}
\usepackage{enumitem}
\usepackage{cancel}
\usepackage{tcolorbox}
\usepackage{soul} %

\usepackage{etoolbox}
\appto\appendix{\addtocontents{toc}{\protect\setcounter{tocdepth}{1}}}
\appto\listoffigures{\addtocontents{lof}{\protect\setcounter{tocdepth}{1}}}
\appto\listoftables{\addtocontents{lot}{\protect\setcounter{tocdepth}{1}}}

\usepackage[ruled,linesnumbered]{algorithm2e}
\SetAlgorithmName{Pseudocode}{pscode}{List of Pseudocodes}

\newcommand{\dt}{{\Delta{t}}}

\def\subsubsection{\subsection}

\DeclareRobustCommand{\[}{\begin{equation}}
\DeclareRobustCommand{\]}{\end{equation}}

\let\Tuftemarginnote\marginnote
\let\marginnote\relax
\usepackage{marginnote}

\let\marginnote\Tuftemarginnote

\def\mathnote#1{%
  \tag*{\rlap{\hspace\marginparsep{\parbox[t]{\marginparwidth}{\footnotesize#1}}}}%
}

\usepackage[hyperpageref]{backref}
\definecolor{mydarkblue}{rgb}{0,0.08,0.45}
\hypersetup{ %
    colorlinks=true,
    linkcolor=mydarkblue,
    citecolor=mydarkblue,
    filecolor=mydarkblue,
    urlcolor=mydarkblue,
    pdfview=FitH,
}

\newcommand{\half}{\nicefrac{1}{2}}

\DeclareMathOperator*{\argmin}{argmin}

\newcommand{\R}{\mathbb{R}}

\newcommand{\E}{\mathop{\mathbb{E}}}
\renewcommand{\bar}{\overline}
\renewcommand{\epsilon}{\varepsilon}
\newcommand{\eps}{\varepsilon}

\newcommand{\flowto}[1]{\overset{#1}{\hookrightarrow} }
\newcommand{\goto}[1]{\overset{#1}{\longrightarrow} }

\newcommand{\Tab}[2]{v^{[#1,#2]}}
\newcommand{\rflow}{\mathrm{RunFlow}}

\newcommand{\cO}{\mathcal{O}}
\newcommand{\cN}{\mathcal{N}}

\renewcommand{\hat}{\widehat}

\newtheorem{lemma}{Lemma}

\newtheorem{claim}{Claim}
\newtheorem{definition}{Definition}

\newtheorem{fact}{Fact}

\usepackage{silence}
\newcommand{\wip}[1]{{}} %

\newcommand\Mark[1]{\textsuperscript{#1}}

\title{Step-by-Step Diffusion: An Elementary Tutorial}

\author{Preetum Nakkiran\Mark{1}, Arwen Bradley\Mark{1}, Hattie Zhou\Mark{{1,2}}, Madhu Advani\Mark{1}}
\date{\Mark{1}Apple, \Mark{2}Mila, Université de Montréal}

\usepackage{tocloft}

\begin{document}

\maketitle

\begin{abstract}
We present an accessible first course on
diffusion models and flow matching for machine learning,
aimed at a technical audience with no diffusion experience.
We try to simplify the mathematical details as much as possible
(sometimes heuristically), 
while retaining enough precision to
derive correct algorithms.
\end{abstract}

\vspace{-5pt}
\tableofcontents
\newpage

\part*{Preface}

There are many existing resources for learning diffusion models.
Why did we write another?
Our goal was to teach diffusion as simply as possible,
with minimal mathematical and machine learning prerequisites,
but in enough detail to reason about its correctness.
Unlike most tutorials on this subject,
we take neither a Variational Auto Encoder (VAE)
nor an Stochastic Differential Equations (SDE) approach.
In fact, for the core ideas we will not need any SDEs, Evidence-Based-Lower-Bounds (ELBOs), 
Langevin dynamics, or even the notion of a score.
The reader need only be familiar with basic probability, calculus, linear algebra,
and multivariate Gaussians.
The intended audience for this tutorial is
technical readers at the
level of at least advanced undergraduate or graduate students,
who are learning diffusion for the first time
and want a mathematical understanding of the subject.

This tutorial has five parts, each relatively self-contained, but covering closely related topics.
Section \ref{sec:fundamentals} presents the fundamentals of diffusion: the problem we are trying to solve and an overview of the basic approach. Sections \ref{sec:ddpm} and \ref{sec:ddim} show how to construct a stochastic and deterministic diffusion sampler, respectively, and give
intuitive derivations for why
these samplers correctly reverse the forward diffusion process.
Section \ref{sec:flows} covers the closely-related topic of Flow Matching,
which can be thought of as a generalization of diffusion that offers additional flexibility (including what are called rectified flows or linear flows).
Finally, in Section \ref{sec:practical} we return to diffusion and connect this tutorial to the broader literature while highlighting some of the design choices that matter most in practice, including samplers, noise schedules, and parametrizations.

\subsubsection*{Acknowledgements}
We are grateful for helpful feedback and suggestions
from many people, in particular:
Josh Susskind,
Eugene Ndiaye,
Dan Busbridge,
Sam Power,
De Wang,
Russ Webb,
Sitan Chen,
Vimal Thilak,
Etai Littwin,
Chenyang Yuan,
Alex Schwing,
Miguel Angel Bautista Martin,
and Dilip Krishnan.

\newpage

\section{Fundamentals of Diffusion}
\label{sec:fundamentals}

\newthought{The goal} of generative modeling is:
given i.i.d.
samples from some unknown distribution $p^*(x)$,
construct a sampler for (approximately) the same distribution.
For example,
given a training set of dog images from some underlying distribution $p_{\textrm{dog}}$,
we want a method of producing new images of dogs from this distribution.

One way to solve this problem, at a high level,
is to learn a transformation from 
some easy-to-sample distribution (such as Gaussian noise)
to our target distribution $p^*$.
Diffusion models offer a general framework for learning such transformations.
The clever trick of diffusion is to reduce the
problem of sampling from distribution $p^*(x)$ into 
to a sequence of \emph{easier} sampling problems.

This idea is best explained via the following Gaussian diffusion example.
We'll sketch the main ideas now, and in later
sections we will use this setup to derive what are commonly known as the
\emph{DDPM} and \emph{DDIM} samplers\footnote{These stand for Denoising Diffusion Probabilistic Models (DDPM) and Denoising Diffusion Implicit Models (DDIM),
following \citet{ho2020denoising} and \citet{song2021denoising}.}, and reason about their correctness.

\subsection{Gaussian Diffusion}
For Gaussian diffusion, let $x_0$ be
a random variable in $\R^d$ distributed according
to the target distribution $p^*$ (e.g., images of dogs).
Then construct a sequence of random variables $x_1, x_2, \dots, x_T$,
by successively adding independent Gaussian noise with some small scale $\sigma$:
\begin{equation}
\label{eqn:intro-fwd}
    x_{t+1} := x_{t} + \eta_{t} ,\quad \eta_{t} \sim  \cN(0, \sigma^2).
\end{equation}
This is called the \emph{forward process}\footnote{
One benefit of using this particular forward process
is computational: we can directly sample $x_t$ given $x_0$ in constant time.},
which transforms the data distribution into a noise distribution.
Equation~\eqref{eqn:intro-fwd} defines a joint distribution over all $(x_0, x_1, \dots, x_T)$,
and we let $\{p_t\}_{t \in [T]}$ denote the marginal distributions of each $x_t$.
Notice that at large step count $T$,
the distribution $p_T$ is nearly Gaussian\footnote{Formally,
$p_T$ is close in KL divergence to $\cN(0, T\sigma^2)$,
assuming $p_0$ has bounded moments.},
so we can approximately sample from $p_T$ by just sampling a Gaussian.

\begin{figure}
    \classiccaptionstyle
    \centering
    \includegraphics[width=0.95\linewidth]{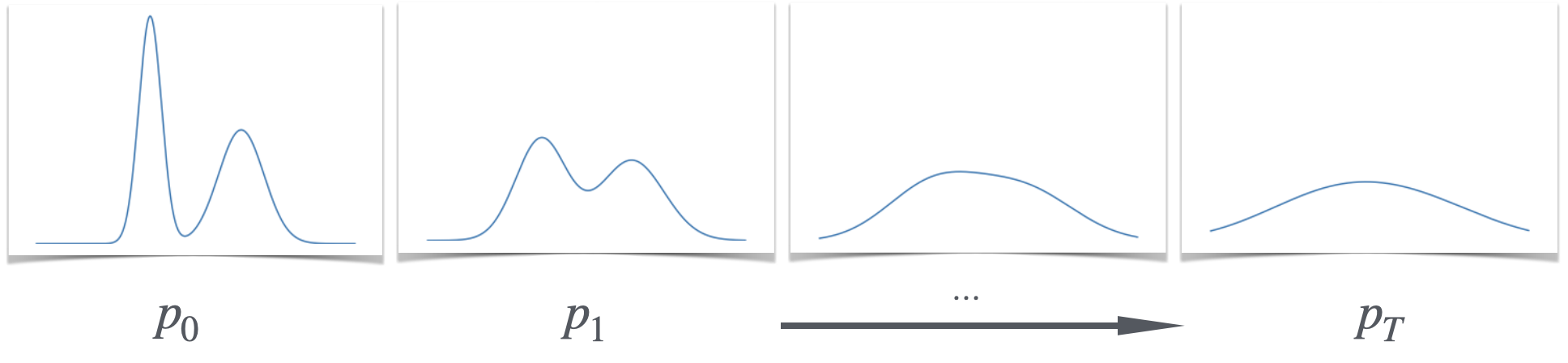}
    \caption{Probability distributions defined by diffusion forward process on one-dimensional target distribution $p_0$.}
    \label{fig:intro_diff}
\end{figure}

\newpage
\noindent
Now, suppose we can solve the following subproblem:
\vspace{-1pt}
\begin{quote}
\emph{
``Given a sample marginally distributed as $p_{t}$,
produce a sample marginally distributed as $p_{t-1}$''}.
\end{quote}
\vspace{-1pt}
We will call a method that does this a \emph{reverse sampler}\footnote{Reverse samplers will be formally defined in Section~\ref{sec:abstract} below.},
since it tells us how to sample from $p_{t-1}$ assuming we can already sample from $p_{t}$.
If we had a reverse sampler,
we could sample from our target $p_0$ by simply starting with
a Gaussian sample from $p_T$, and iteratively applying the
reverse sampling procedure
to get samples from $p_{T-1}, p_{T-2}, \dots$ and finally $p_0=p^*$.

The key insight of diffusion is,
\textbf{learning to reverse each intermediate step
can be easier}
than learning to 
sample from the target distribution in one step\footnote{
Intuitively this is because the distributions
$(p_{t-1}, p_{t})$ are already quite close,
so the reverse sampler does not need to do much.}.
There are many ways to construct reverse samplers,
but for concreteness let us first see the standard diffusion sampler
which we will call the \emph{DDPM sampler}\footnote{This is the sampling strategy originally proposed in \citet{OGdiffusion}.}.

The \emph{Ideal DDPM sampler} uses the obvious strategy:
    At time $t$,
    given input $z$
    (which is promised to be a sample from $p_t$),
    we output a sample from the
    conditional distribution
   \vspace{-5pt}
   \[
   \label{eqn:ideal_ddpm_sampler}
   p(x_{t-1} \mid x_{t} = z).
   \vspace{-3pt}
   \]
This is clearly a correct reverse sampler. The problem is,
it requires learning a
generative model for the
conditional distribution
${p(x_{t-1} \mid x_{t})}$ for every $x_{t}$, which
could be complicated.
But if the per-step noise $\sigma$ is sufficiently small,
then it turns out this conditional distribution
becomes simple:

\begin{figure*}[b]
    \classiccaptionstyle
    \centering
    \includegraphics[width=0.98\linewidth]{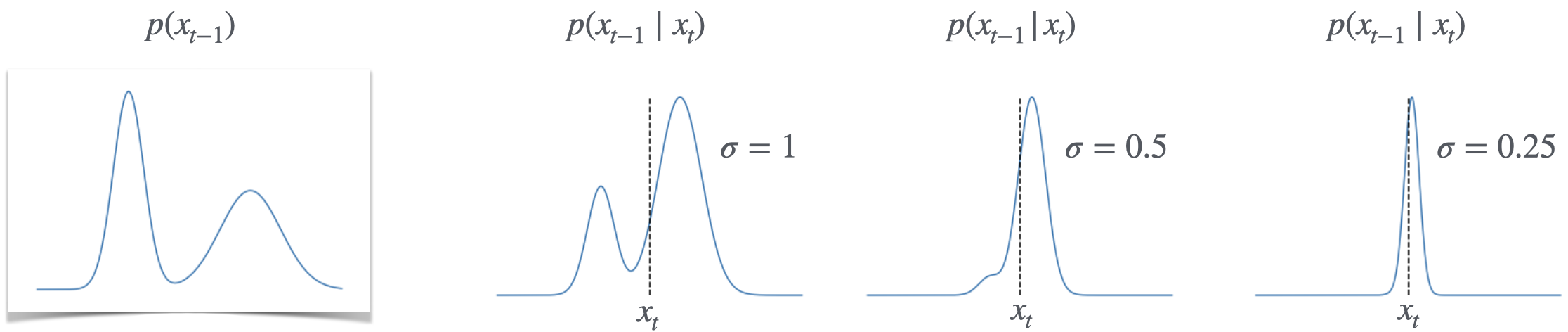}
    \caption{Illustration of Fact 1.
    The prior distribution \mbox{$p(x_{t-1})$}, leftmost,
    defines a joint distribution $(x_{t-1}, x_t)$
    where \mbox{$p(x_t \mid x_{t-1}) = \cN(0, \sigma^2)$}.
    We plot the reverse conditional distributions
    \mbox{$p(x_{t-1} \mid x_t)$}
    for a fixed conditioning $x_t$,
    and varying noise levels $\sigma$.
    Notice these distributions 
    become close to Gaussian for small $\sigma$.
    }
    \label{fig:post_diff}
\end{figure*}

\begin{fact}[Diffusion Reverse Process]

\label{fact:rev_diff}
For small $\sigma$,
and the Gaussian diffusion process defined in~\eqref{eqn:intro-fwd},
the conditional distribution $p(x_{t-1} \mid x_{t})$ 
is itself close to Gaussian.
That is, for all times $t$ and conditionings $z \in \R^d$,
there exists some mean parameter $\mu \in \R^d$
such that
\vspace{-5pt}
\begin{equation}
\label{eqn:gauss_ddpm}
   p(x_{t-1} \mid x_{t} = z)
   \approx \cN(x_{t-1} ; ~~\mu ~,~ \sigma^2).
\end{equation}
\end{fact}
This is not an obvious fact; we will derive it in Section~\ref{sec:ddpm_valid}.
This fact
enables a drastic simplification:
instead of having to learn
an arbitrary distribution $p(x_{t-1} \mid x_{t})$
from scratch,
we now know everything about this distribution
except its mean, which we denote\footnote{
We denote the mean as a function $\mu_{t-1}: \R^d \to \R^d$
because the mean of $p( x_{t-1} \mid x_t )$
depends on the time $t$ as well as the conditioning
$x_{t}$, as described in Fact~\ref{fact:rev_diff}.
} $\mu_{t-1}(x_{t})$. The fact that we can approximate the posterior distribution as Gaussian when $\sigma$ is sufficiently small is illustrated in Fig \ref{fig:post_diff}. 
This is an important point, so to re-iterate:
for a given time $t$ and conditioning value $x_{t}$,
learning the mean of
$p( x_{t-1} \mid x_{t})$
is sufficient to learn the full conditional distribution
$p( x_{t-1} \mid x_{t})$.

Learning the mean of $p(x_{t-1} \mid x_t)$
is a much simpler problem than learning the full conditional distribution, because we can solve it by regression. 
To elaborate, we have a joint distribution
$(x_{t-1}, x_{t})$ from which we can easily sample,
and we would like to estimate $\E[ x_{t-1} \mid x_{t}]$.
This can be done by optimizing a standard regression loss\footnote{
Recall the generic fact that for any distribution over
$(x, y)$, we have: $\argmin_{f} \E ||f(x) - y||^2 = \E[y \mid x]$
}:
\begin{align}
{\mu}_{t-1}(z) 
&:= \E[ x_{t-1} \mid x_{t} = z]\\
\implies \mu_{t-1} &= \argmin_{f: \R^d \to \R^d}
~~\E_{x_t, x_{t-1}} || f(x_{t}) - x_{t-1} ||_2^2 \\
&= \argmin_{f: \R^d \to \R^d}
\E_{x_{t-1},\eta}
|| f(x_{t-1} + \eta_t) - x_{t-1}) ||_2^2,
\label{eqn:denoising}
\end{align}
where the expectation
is taken over samples
$x_0$ from our target distribution $p^*$.\footnote{Notice that we simulate
samples of $(x_{t-1}, x_{t})$ 
by adding noise to the samples of $x_0$, as defined in Equation~\ref{eqn:intro-fwd}.}
This particular regression problem is well-studied in certain settings.
For example, when the target $p^*$ is a distribution on images,
then the corresponding regression problem (Equation~\ref{eqn:denoising}) is exactly an
\emph{image denoising objective}, which can be approached with familiar methods (e.g. convolutional neural networks).

\newthought{Stepping back}, we have seen something remarkable:
we have reduced the problem of learning
to sample from an arbitrary distribution to 
the standard problem of regression.

\subsection{Diffusions in the Abstract}
\label{sec:abstract}
Let us now abstract away the Gaussian setting,
to define diffusion-like models in a way that will capture
their many instantiations (including deterministic samplers,
discrete domains, and flow-matching).

Abstractly, here is how to construct a diffusion-like generative model:
We start with our target distribution $p^*$,
and we pick some base distribution $q(x)$
which is easy to sample from,
e.g. a standard Gaussian or i.i.d bits.
We then try to construct a sequence of distributions 
which interpolate between our target $p^*$ and the base distribution $q$.
That is,
we construct distributions
\[
p_0 ~,~ p_1 ~,~ p_2 ~,~ \dots,  p_T,
\]
such that $p_0 = p^*$ is our target, $p_T = q$ the base distribution,
and adjacent distributions $(p_{t-1}, p_{t})$ are marginally ``close'' 
in some appropriate sense.
Then, we learn a \emph{reverse sampler} which transforms distributions
$p_{t}$ to $p_{t-1}$. This is the key learning step, which presumably
is made easier by the fact that adjacent distributions are ``close.''
Formally, reverse samplers are defined below.
\begin{definition}[Reverse Sampler]
\label{def:rsamp}
Given a sequence of marginal distributions $p_t$,
a \emph{reverse sampler}
for step $t$ is a potentially stochastic function
$F_t$
such that
if $x_{t} \sim p_{t}$, then
the marginal distribution of $F_t(x_{t})$
is exactly $p_{t-1}$:
\[
\{ F_t(z) : z \sim p_t \} \equiv p_{t-1}.
\]
\end{definition}

There are many possible reverse samplers\footnote{Notice that none of this abstraction is specific to the case of Gaussian noise--- in fact, it does not even require the concept of ``adding noise''.
It is even possible to instantiate in discrete settings,
where we consider distributions $p^*$ over a finite set,
and define corresponding ``interpolating distributions''
and reverse samplers.},
and it is even possible
to construct reverse samplers
which are \emph{deterministic}.
In the remainder of this tutorial we will
see three popular reverse samplers more formally:
the \emph{DDPM} sampler discussed above (Section~\ref{sec:ddpm_valid}),
the \emph{DDIM} sampler (Section~\ref{sec:ddim}),
which is deterministic,
and the family of \emph{flow-matching models} (Section~\ref{sec:flows}),
which can be thought of as a generalization of DDIM.\footnote{
Given a set of marginal distributions $\{p_t\}$,
there are many possible joint distributions consistent with these marginals
(such joint distributions are called \emph{couplings}).
There is therefore no canonical reverse sampler for a given set of marginals $\{p_t\}$ --- we are free
to chose whichever coupling is most convenient.
}

\subsection{Discretization}
\label{sec:discretization}

Before we proceed further, we need to be more precise about what we mean by adjacent distributions $p_{t}, p_{t-1}$ being ``close". We want to think of the sequence $p_0, p_1, \ldots, p_T$ as the discretization of some (well-behaved) time-evolving function $p(x, t)$, that starts from the target distribution $p_0$ at time $t=0$ and ends at the
noisy distribution $p_T$ at time $t=1$:
\[
\label{eqn:fwd_discretized}
p(x, k \dt) = p_k(x), \quad \text{where } \dt = \frac{1}{T}.
\]
The number of steps $T$ controls the fineness of the discretization (hence the closeness of adjacent distributions).\footnote{This naturally suggests taking the continuous-time limit, which we discuss in Section~\ref{sec:sdes}, though it is not needed for most of our arguments.}

In order to ensure that the variance of the final distribution, $p_T$, is independent of the number
of discretization steps, we also need to be more specific about the variance of each increment.
Note that if $x_k = x_{k-1} + \cN(0, \sigma^2)$, then $x_T \sim \cN(x_0, T \sigma^2)$. Therefore, we need to scale the variance of each increment by $\dt = 1/T$, that is, choose
\begin{align}
    \sigma = \sigma_q \sqrt{\dt},
\end{align}
where $\sigma_q^2$ is the desired terminal variance.
This choice ensures that the variance of $p_T$ is always $\sigma_q^2$, regardless of $T$. (The $\sqrt{\dt}$ scaling will turn out to be important in our arguments for the correctness of our reverse solvers in the next chapter, and also connects to the SDE formulation in Section \ref{sec:sdes}.)

At this point, it is convenient to adjust our notation. From here on, $t$ will represent a continuous-value in the interval $[0,1]$ (specifically, taking one of the values $0, \dt, 2\dt, \ldots, T \dt=1$). Subscripts will indicate \emph{time} rather than \emph{index}, so for example $x_t$ will now denote $x$ at a discretized time $t$. That is, Equation \ref{eqn:intro-fwd} becomes:
\begin{equation}
\label{eqn:intro-fwd-dt}
    x_{t+\dt} := x_{t} + \eta_{t}, \quad \eta_{t} \sim  \cN(0, \sigma_q^2 \dt), 
\end{equation}
which also implies that
\begin{equation}
\label{eqn:intro-xt-dt}
    x_{t} \sim  \cN(x_0, \sigma_t^2), \quad
    \text{where } \sigma_t := \sigma_q \sqrt{t},
\end{equation}
since the total noise added up to time $t$
(i.e. $\sum_{\tau \in \{0, \dt, 2\dt, \dots, t-\dt\}} \eta_\tau$)
is also Gaussian with mean zero
and variance $\sum_\tau \sigma_q^2 \dt = \sigma_q^2 t$.

\clearpage
\newpage
\section{Stochastic Sampling: DDPM}
\label{sec:ddpm}

In this section we review
the DDPM-like
reverse sampler discussed in Section~\ref{sec:fundamentals},
and heuristically prove its correctness.
This sampler is conceptually the same
as the sampler popularized in \emph{Denoising Diffusion Probabilistic Models} (DDPM) by \citet{ho2020denoising}
and originally introduced by \citet{OGdiffusion}, when adapted
to our simplified setting.
However, a word of warning for the reader familiar with \citet{ho2020denoising}:
Although the overall strategy of our sampler is identical to \citet{ho2020denoising},
certain technical details (like constants, etc) are 
slightly different\footnote{\label{foot:ddpm_diff}
For the experts, the main difference is we use the ``Variance Exploding''
diffusion forward process. We also use a constant noise schedule, and
we do not discuss how to parameterize the predictor (``predicting $x_0$ vs. $x_{t-1}$ vs. noise $\eta$'').
We elaborate on the latter point in Section~\ref{sec:predict_x0}.
}.

We consider the setup from Section~\ref{sec:discretization},
with some target distribution $p^*$
and the joint distribution of noisy samples $(x_0, x_\dt, \dots, x_1)$
defined by Equation~\eqref{eqn:intro-fwd-dt}.
The DDPM sampler will require estimates of the following conditional expectations:
\begin{align}
    \mu_{t}(z) := \E[x_{t} \mid x_{t+\dt} = z].
\end{align}
This is a set of functions $\{\mu_t\}$, one for every time step
$t \in \{0, \dt, \dots, 1-\dt\}$.
In the \emph{training} phase, we 
estimate these functions
from i.i.d. samples of $x_0$,
by optimizing the denoising regression objective
\begin{align}
\mu_{t} &= \argmin_{f: \R^d \to \R^d}
~~\E_{x_t, x_{t+\dt}} || f(x_{t+\dt}) - x_{t} ||_2^2 ~,
\label{eqn:denoising-dt}
\end{align}
typically with a neural-network\footnote{
In practice, it is common to share parameters 
when learning the different regression functions $\{\mu_t\}_t$,
instead of learning a separate function for each timestep independently.
This is usually implemented by training a model $f_\theta$
that accepts the time $t$ as an additional argument,
such that $f_\theta(x_t, t) \approx \mu_t(x_t)$.}
parameterizing $f$.
Then, in the \emph{inference} phase, we use the estimated functions in the following
reverse sampler.

\begin{mdframed}[nobreak=true]
\underline{Algorithm 1: Stochastic Reverse Sampler (DDPM-like)}\\
For input sample $x_t$, and timestep $t$, output:
\begin{align}
\label{eqn:ddpm}
    \hat{x}_{t-\dt} \gets {\mu}_{t-\dt}(x_t) + \cN(0, \sigma^2_q \dt)
\end{align}
\end{mdframed}
To actually generate a sample, we first sample $x_1$
as an isotropic Gaussian $x_1 \sim \cN(0, \sigma_q^2)$,
and then run the iteration of Algorithm 1 down to $t=0$,
to produce a generated sample $\hat{x}_0$. (Recall that in our discretized notation~\eqref{eqn:intro-xt-dt}, $x_1$ is the fully-noised terminal distribution, and the iteration takes steps of size $\dt$.)
Explicit pseudocode for these algorithms are given in Section~\ref{sec:ddpm_algos}.

We want to reason about correctness of this entire procedure:
why does iterating Algorithm 1 produce a sample from [approximately] our target distribution $p^*$?
The key missing piece is, we need to prove some version of Fact~\ref{fact:rev_diff}:
that the true conditional $p(x_{t-\dt} \mid x_t)$ can be 
well-approximated by a Gaussian, and this approximation
gets better as we scale $\dt \to 0$.

\subsection{Correctness of DDPM}
\label{sec:ddpm_valid}

Here is a more precise version of Fact~\ref{fact:rev_diff},
along with a heuristic derivation.
This will complete the argument that Algorithm 1 is correct--- i.e.
that it approximates a valid reverse sampler in the sense of Definition~\ref{def:rsamp}.

\begin{claim}[Informal]
\label{claim:ddpm_main}
Let $p_{t-\dt}(x)$ be an arbitrary, sufficiently-smooth density over $\R^d$.
Consider the joint distribution of $(x_{t-\dt}, x_{t})$,
where $x_{t-\dt} \sim p_{t-\dt}$ and
$x_{t} \sim x_{t-\dt} + \cN(0, \sigma_q^2 \dt )$.
Then, for sufficiently small $\dt$, the following holds.
For all conditionings $z \in \R^d$,
there exists $\mu_z$ such that:
\begin{align}
\label{eqn:approx_fact}
p(x_{t-\dt} \mid x_{t} = z) \approx \cN(x_{t-\dt} ; ~~\mu_z ~,~ \sigma_q^2 \dt).
\end{align}
for some constant $\mu_z$ depending only on $z$.
Moreover, it suffices to take\footnote{Experts will recognize this mean as related to the \emph{score}.
In fact, Tweedie's formula implies that this mean is exactly correct even for large $\dt$,
with no approximation required. That is,
$\E[ x_{t-\dt} \mid x_t=z ] = z + \sigma_q^2 \dt \nabla \log p_t(z)$.
The distribution $p(x_{t-\dt} \mid x_t)$ may deviate from Gaussian,
however, for larger $\sigma$.}
\begin{align}
\mu_z &:= \E_{(x_{t-\dt}, x_t)}[x_{t-\dt} \mid x_{t} = z] \\
&= z + (\sigma_q^2 \dt) \nabla \log p_{t}(z), \label{eqn:score}
\end{align}
where $p_t$ is the marginal distribution of $x_t$.
\end{claim}

Before we see the derivation, a few remarks:
Claim~\ref{claim:ddpm_main} implies that to sample from $x_{t-\dt}$,
it suffices to first sample from $x_t$, then sample from 
a \emph{Gaussian} distribution centered around $\E[x_{t-\dt} \mid x_t]$.
This is exactly what DDPM does, in Equation~\eqref{eqn:ddpm}.
Finally, in these notes we will not actually
need the expression for $\mu_z$ in Equation~\eqref{eqn:score};
it is enough for us know that such a $\mu_z$ exists,
so we can learn it from samples.

\begin{proof}[Proof of Claim ~\ref{claim:ddpm_main} (Informal)]
Here is a heuristic argument for why the score appears in the reverse process.
We will essentially just apply Bayes rule and then Taylor expand appropriately.
We start with Bayes rule:
\begin{align}
p(x_{t-\dt} | x_{t}) &= p( x_{t} | x_{t-\dt} ) p_{t-\dt}(x_{t-\dt}) / p_{t}(x_{t})
\end{align}
Then take logs of both sizes.
Throughout, we will drop any additive constants in the log (which translate to normalizing factors),
and drop all terms of order $\mathcal{O}(\dt)$ \footnote[][-1cm]{Note that $x_{t+1} - x_{t} \sim \mathcal{O}(\sqrt{\dt})$. Dropping $\mathcal{O}(\dt)$ terms means dropping $(x_{t+1} - x_{t})^2 \sim \mathcal{O}(\dt)$ in the expansion of $p_t(x_t)$, but keeping $\frac{1}{2 \sigma_q^2 \dt}(x_{t+1} - x_{t})^2 \sim \mathcal{O}(1)$ in $p(x_{t} | x_{t+1})$.}.
Note that we should think of $x_{t}$ as a constant in this derivation, since we want to 
understand the conditional probability as a function of $x_{t-\dt}$. Now: 
\begin{align*}
&\log p(x_{t-\dt} | x_{t}) = \log p(x_{t} | x_{t-\dt} ) + \log p_{t-\dt}(x_{t-\dt})  \cancel{-\log p_t(x_t)}
\mathnote{Drop constants involving only $x_{t}$.}\\
&= \log p(x_{t} | x_{t-\dt} ) + \log p_{t}(x_{t-\dt}) + \mathcal{O}(\dt) \mathnote{Since $p_{t-\dt}(\cdot) = p_t(\cdot) + \dt \frac{\partial}{\partial t} p_t(\cdot)$.}\\
&= -\frac{1}{2 \sigma_q^2 \dt} ||x_{t-\dt} -  x_{t}||_2^2 + \log p_{t}(x_{t-\dt}) \mathnote{Definition of $\log p(x_{t} | x_{t-\dt} )$.}\\
&= -\frac{1}{2 \sigma_q^2 \dt} ||x_{t-\dt} -  x_{t}||_2^2  \\
&\quad\quad + \cancel{\log p_{t}(x_t)}
+ \langle \nabla_x \log p_{t}(x_{t}), (x_{t-\dt} - x_{t})\rangle +\mathcal{O}(\dt) 
\mathnote{Taylor expand around $x_t$ and drop constants.} \\
&= -\frac{1}{2 \sigma_q^2 \dt} \left(
||x_{t-\dt} -  x_{t}||_2^2 - 2\sigma_q^2 \dt \langle \nabla_x \log p_{t}(x_{t}), (x_{t-\dt} - x_{t})\rangle 
\right) \\
&= -\frac{1}{2 \sigma_q^2 \dt}
||x_{t-\dt} -  x_{t} - \sigma_q^2 \dt ~\nabla_x \log p_{t}(x_{t})||_2^2  + C
\mathnote{Complete the square in \mbox{$(x_{t-\dt}-x_{t})$}, and drop constant $C$ involving only $x_{t}$.}\\
&= -\frac{1}{2 \sigma_q^2 \dt}
||x_{t-\dt} - \mu||_2^2 
\mathnote{For $\mu := x_t + (\sigma_q^2 \dt)\nabla_x \log p_t(x_{t})$.}
\end{align*}
This is identical, up to additive factors, to the log-density of a Normal distribution with mean $\mu$
and variance $\sigma_q^2 \dt$. Therefore,
\[
p(x_{t-\dt} \mid x_{t}) \approx \cN(x_{t-\dt} ;~ \mu, \sigma_q^2 \dt).
\]
\end{proof}

Reflecting on this derivation,
the main idea was that for small enough $\dt$,
the Bayes-rule expansion of the reverse process $p(x_{t-\dt} \mid x_t)$
is dominated by the term $p(x_t \mid x_{t-\dt})$, from the forward process.
This is intuitively why the reverse process and the forward process
have the same functional form (both are Gaussian here)\footnote{This general relationship 
between forward
and reverse processes holds somewhat more generally
than just Gaussian diffusion; see e.g. the discussion in \citet{OGdiffusion}.}.

\paragraph{Technical Details [Optional].}
The meticulous reader may notice that Claim~\ref{claim:ddpm_main}
is not obviously sufficient to imply correctness of the entire DDPM algorithm.
The issue is: as we scale down $\dt$,
the error in our per-step approximation (Equation~\ref{eqn:approx_fact}) decreases,
but the number of total steps required increases.
So if the per-step error does not decrease fast enough (as a function of $\dt$),
then these errors could accumulate to a non-negligible error by the final step.
Thus, we need to quantify how fast the per-step error decays.
Lemma~\ref{lem:KL} below is one way of quantifying this:
it states that if the step-size (i.e. variance of the per-step noise) is $\sigma^2$,
then the KL error of the per-step Gaussian approximation is $\cO(\sigma^4)$.
This decay rate is fast enough,
because the number of steps only grows as\footnote{
The chain rule for KL implies that we can add up these per-step errors:
the approximation error for the final sample is bounded by the sum of
all the per-step errors.
} $\Omega(1/\sigma^2)$.

\begin{lemma}
\label{lem:KL}
Let $p(x)$ be an arbitrary density over $\R$,
with bounded 1st to 4th order derivatives.
Consider the joint distribution $(x_0, x_1)$,
where $x_0 \sim p$ and $x_1 \sim x_0 + \cN(0, \sigma^2)$.
Then, for any conditioning $z \in \R$,
we have
\[
\mathrm{KL}\left(
\cN(\mu_z, \sigma^2 ) 
~~||~~
p_{x_0 \mid x_1}(\cdot \mid x_1 = z)
\right)
\leq O(\sigma^4),
\]
where
\[
\mu_z := z + \sigma^2 \nabla \log p(z).
\]
\end{lemma}
It is possible to prove Lemma~\ref{lem:KL} by doing essentially a careful Taylor expansion;
we include the full proof in Appendix~\ref{app:kl}.

\subsection{Algorithms}
\label{sec:ddpm_algos}

Pseudocode listings \ref{alg:ddpm_train} and \ref{alg:ddpm_gen}
give the explicit DDPM train loss
and sampling code.
To train\footnote{Note that the training procedure
optimizes $f_\theta$ for
all timesteps $t$ simultaneously, by sampling $t \in [0, 1]$ uniformly
in Line 2.} the network $f_\theta$,
we must minimize the expected loss $L_\theta$
output by Pseudocode~\ref{alg:ddpm_train},
typically by backpropagation.

Pseudocode~\ref{alg:ddim_inf} describes the closely-related \emph{DDIM} sampler,
which will be discussed later in Section~\ref{sec:ddim}.

\begin{fullwidth}

\begin{minipage}[t]{0.47\linewidth}
\begin{algorithm}[H]
\DontPrintSemicolon
\caption{DDPM train loss} \label{alg:ddpm_train}
\KwIn{Neural network $f_\theta$;
Sample-access to target distribution $p$.}
\KwData{Terminal variance $\sigma_q$; step-size $\dt$.}
\KwOut{Stochastic loss $L_\theta$}
$x_0 \gets \textrm{Sample}(p)$ \;
$t \gets \textrm{Unif}[0, 1]$ \;
$x_{t} \gets x_0 + \cN(0, \sigma_q^2 t) $ \;
$x_{t+\dt} \gets x_{t} + \cN(0, \sigma_q^2 \dt) $ \;
$L \gets \left\Vert f_\theta(x_{t+\dt}, t+\dt) -  x_{t}\right\Vert_2^2$ \;
\Return $L_\theta$
\end{algorithm}
\end{minipage}
\hfill
\begin{minipage}[t]{0.47\linewidth}
\begin{algorithm}[H]
\DontPrintSemicolon
\caption{DDPM sampling (Code for Algorithm 1)}\label{alg:ddpm_gen}
\KwIn{Trained model $f_\theta$.}
\KwData{Terminal variance $\sigma_q$; step-size $\dt$.}
\KwOut{$x_0$}
$x_1 \gets \cN(0, \sigma_q^2)$ \;
\For{$t=1,~(1-\dt),(1-2\dt), \dots, \dt$}{
$\eta \gets \cN(0, \sigma_q^2 \dt)$  \;
$x_{t-\dt} \gets f_\theta(x_t, t) + \eta$ \;
}
\Return $x_0$
\end{algorithm}
\end{minipage}

\begin{algorithm}
\DontPrintSemicolon
\caption{DDIM sampling (Code for Algorithm 2)}\label{alg:ddim_inf}
\KwIn{Trained model $f_\theta$}
\KwData{Terminal variance $\sigma_q$; step-size $\dt$.}
\KwOut{$x_0$}
$x_1 \gets \cN(0, \sigma_q^2)$ \;
\For{$t=1,~(1-\dt),(1-2\dt), \dots, \dt, 0$}{
$\lambda \gets \frac{\sqrt{t}}{\sqrt{t-\dt} + \sqrt{t}}$ \;
$x_{t-\dt} \gets x_t + \lambda(f_\theta(x_t, t) - x_t)$ \;
}
\Return $x_0$
\end{algorithm}

\end{fullwidth}

\subsection{Variance Reduction: Predicting $x_0$}
\label{sec:predict_x0}

Thus far, our diffusion models have been
trained to predict \mbox{$\E[x_{t-\dt} \mid x_t]$}:
this is what Algorithm 1 requires,
and what the training procedure of Pseudocode~\ref{alg:ddpm_train} produces.
However, many practical diffusion implementations actually 
train to predict $\E[ x_0 \mid x_t ]$, i.e. to predict the expectation of the initial point $x_0$ instead of the previous point $x_{t-\dt}$.
This difference turns out to be just a \emph{variance reduction}
trick, which estimates the same quantity in expectation.
Formally, the two quantities can be related as follows:

\begin{claim}
\label{claim:var_red}
For the Gaussian diffusion setting of Section~\ref{sec:discretization} , we have:
\begin{align}
\E[(x_{t-\dt} - x_t) \mid x_t]  
= 
\frac{\dt}{t}
\E[(x_0 - x_t) \mid x_t]. \label{eq:var_reduc_dt}
\end{align} 
Or equivalently:
\[
\E[x_{t-\dt} \mid x_t]  
= 
\left(\frac{\dt}{t}\right)
\E[x_0  \mid x_t]
+ \left(1- \frac{\dt}{t}\right)x_t.
\label{eqn:var_red}
\]
\end{claim}

This claim implies that if we want to estimate $\E[x_{t-\dt} \mid x_t]$,
we can instead estimate $\E[x_0 \mid x_t]$ and then then essentially divide by $(t/\dt)$,
which is the number of steps taken thus far.
The variance-reduced versions of the DDPM training and sampling algorithms do exactly this;
we include them in Appendix~\ref{app:var-red}.

\begin{marginfigure}
\includegraphics[width=\linewidth]{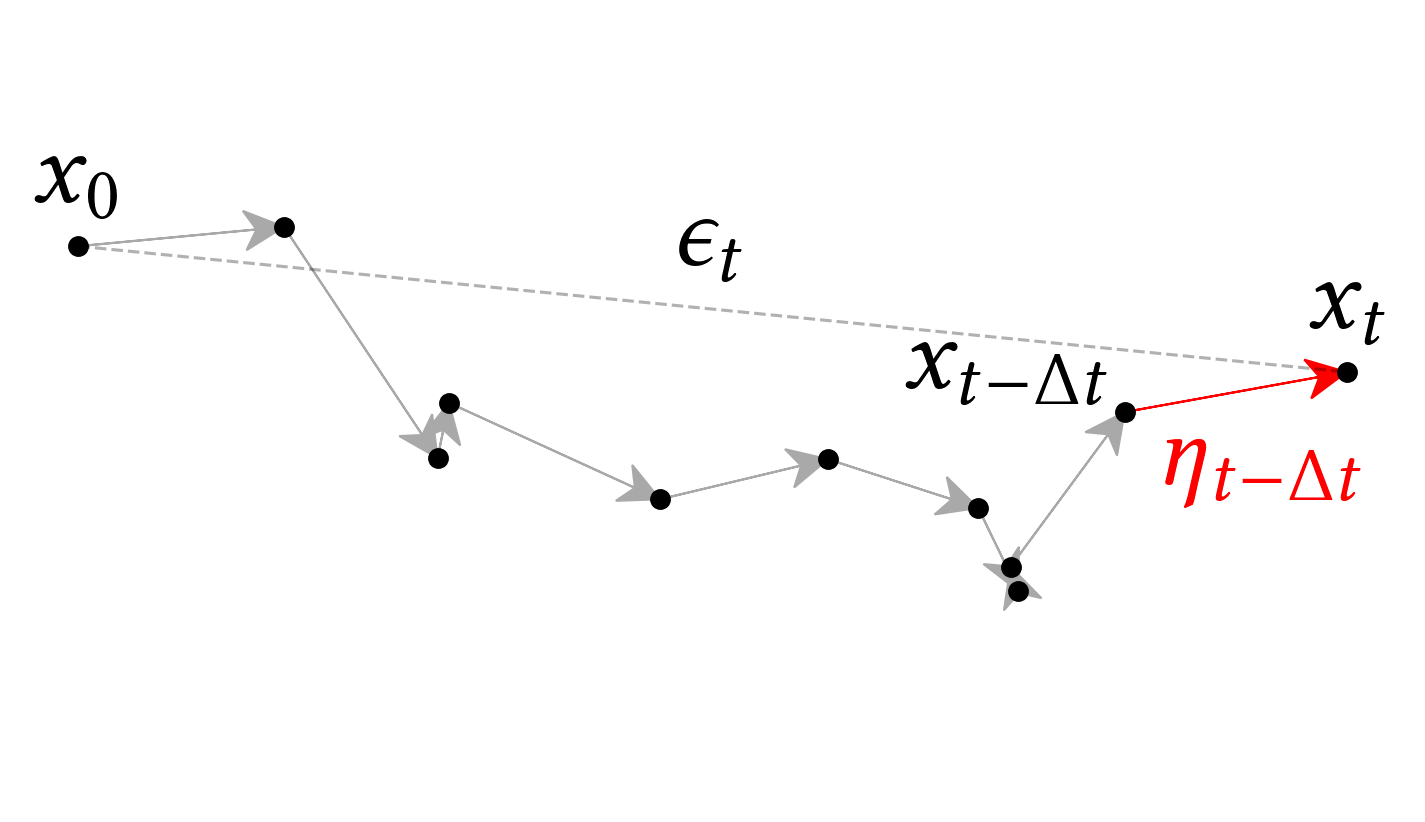}
      \caption{The intuition behind Claim~\ref{claim:var_red}.
      Given $x_t$, the final noise step $\eta_{t-\dt}$
      is distributed identically as all other noise steps,
      intuitively because we only know the sum $x_t = x_0 + \sum_i \eta_i$.
      }
      \label{fig:var-red}
\end{marginfigure}
The intuition behind Claim~\ref{claim:var_red} is illustrated in
Figure~\ref{fig:var-red}: first, 
observe that predicting $x_{t-\dt}$ given $x_t$
is equivalent to predicting the last noise step,
which is $\eta_{t-\dt} = (x_t - x_{t-\dt})$ in the forward process of Equation~\eqref{eqn:intro-fwd-dt}.
But, if we are only given the final $x_{t}$,
then all of the previous noise steps $\{\eta_i\}_{i < t}$ intuitively ``look the same''---
we cannot distinguish between noise that was added at the last step
from noise that was added at the 5th step, for example.
By this symmetry, we can conclude that all of the individual
noise steps are distributed \emph{identically} (though not independently) given $x_t$.
Thus, instead of estimating a single noise step, we can equivalently estimate
the \emph{average of all prior noise steps}, which has much lower variance.
There are $(t/\dt)$ elapsed noise steps by time $t$, so we divide 
the total noise by this quantity in Equation~\ref{eq:var_reduc_dt} to compute the average.
See Appendix~\ref{app:var_proof} for a formal proof.

\newthought{Word of warning:} 
Diffusion models should always be trained to estimate \emph{expectations}.
In particular, when we train a model to predict $\E[x_0 \mid x_t]$,
we should \emph{not} think of this as trying to learn
``how to sample from the distribution $p(x_0 \mid x_t)$''.
For example, if we are training an image diffusion model,
then the optimal model will output $\E[x_0 \mid x_t]$
which will look like a blurry mix of images (e.g. Figure 1b in \citet{karras2022elucidating})--- it will not look like an actual image sample.
It is good to keep in mind that when diffusion papers colloquially discuss
models ``predicting $x_0$'', they do not mean producing something that looks like an actual sample
of $x_0$.

\subsection{Diffusions as SDEs [Optional]}
\label{sec:sdes}

In this section\footnote{Sections marked ``[Optional]'' are advanced material, and can be skipped on first read. None of the main sections depend on Optional material.}, we connect the discrete-time processes we have discussed so far to stochastic differential equations (SDEs). In the continuous limit, as $\dt \to 0$, our discrete diffusion process turns into a stochastic differential equation. SDEs can also represent many other diffusion variants (corresponding to different drift and diffusion terms), offering flexibility in design choices, like scaling and noise-scheduling. The SDE perspective is powerful because existing theory provides a general closed-form solution for the time-reversed SDE. Discretization of the reverse-time SDE for our particular diffusion immediately yields the sampler we derived in this section, but reverse-time SDEs for other diffusion variants are also available \emph{automatically} (and can then be solved with any off-the-shelf or custom SDE solver), enabling better training and sampling strategies as we will discuss further in Section \ref{sec:practical}. Though we mention these connections only briefly here, the SDE perspective has had significant impact on the field. For a more detailed discussion, we recommend Yang Song's blog post \citep{songblog2021generative}.

\subsubsection*{The Limiting SDE}
Recall our discrete update rule:
\begin{align*}
    x_{t+\dt} &= x_{t} + \sigma_q \sqrt{\dt} \xi, \quad \xi \sim \mathcal{N}(0,1).
\end{align*}
In this limit as $\dt \to 0$, this corresponds to a zero-drift SDE:
\begin{align}
    dx &= \sigma_q dw, \label{eq:simple_sde}
\end{align}
where $w$ is a Brownian motion. A Brownian motion is a stochastic process with i.i.d. Gaussian increments whose variance scales with $\dt$.\footnote{See \citet{eldan_blog_sde} for a high-level overview of Brownian motions and It\^{o}’s formula. See also \citet{evans2012introduction} for a gentle introductory textbook, 
and \citet{kloeden2011numerical} for numerical methods.} \emph{Very heuristically}, we can think of 
$dw \sim \lim_{\dt \to 0} \sqrt{\dt} \mathcal{N}(0,1),$ and thus ``derive''~\eqref{eq:simple_sde} by
\begin{align*}
    dx &= \lim_{\dt \to 0} (x_{t+\dt} - x_{t}) = \sigma_q \lim_{\dt \to 0} \sqrt{\dt} \xi  = \sigma_q dw.\\
\end{align*}

More generally, different variants of diffusion are equivalent to SDEs with different choices of drift and diffusion terms:
\begin{align}
    dx &= f(x,t)dt + g(t)dw. \label{eq:general_sde}
\end{align}
The SDE (\ref{eq:simple_sde}) simply has $f = 0$ and $g = \sigma_q$. This formulation encompasses many other possibilities, though, corresponding to different choices of $f$, $g$ in the SDE. As we will revisit in Section \ref{sec:practical}, this flexibility is important for developing effective algorithms. Two important choices made in practice are tuning the noise schedule and scaling $x_t$; together these can help to control the variance of $x_t$, and control how much we focus on different noise levels. Adopting a flexible noise schedule $\{ \sigma_t \}$ in place of the fixed schedule $\sigma_t \equiv \sigma_q \sqrt{t}$ corresponds to the SDE \citep{song2020score}
\begin{align*}
    x_t \sim \cN(x_0, \sigma_t^2) \iff
    x_{t} = x_{t-\dt} + \sqrt{\sigma_t^2 - \sigma_{t-\dt}^2} z_{t-\dt} 
    \iff dx = \sqrt{\frac{d}{dt} \sigma^2(t)} dw. 
\end{align*}
If we also wish to scale each $x_t$ by a factor $s(t)$, \citet{karras2022elucidating} show that this corresponds to the SDE
\footnote{As a sketch of how $f$ arises, let's ignore the noise and note that:
\begin{align*}
    x_t &= s(t) x_0 \\
    \iff x_{t + \dt} &= \frac{s(t+\dt)}{s(t)} x_t \\
    &= x_{t} + \frac{s(t) - s(t + \dt)}{s(t)} x_{t} \\
    \iff dx/dt &= \frac{\dot{s}}{s} x
\end{align*}
}
\begin{align*}
    x_t \sim \cN(s(t) x_0, s(t)^2 \sigma(t)^2) \iff f(x) = \frac{\dot{s}(t)}{s(t)} x, \quad g(t) = s(t) \sqrt{2\dot{\sigma}(t)\sigma(t)}.
\end{align*}
These are only a few examples of the rich and useful design space enabled by the flexible SDE~\eqref{eq:general_sde}.

\subsubsection*{Reverse-Time SDE}
The \emph{time-reversal} of an SDE runs the process \emph{backward in time}. Reverse-time SDEs are the continuous-time analog of samplers like DDPM. A deep result due to \citet{anderson1982reverse} (and nicely re-derived in \citet{winkler_blog_reverse}) states that the time-reversal of SDE~\eqref{eq:general_sde} is given by:
\begin{align}
    dx &= \big( f(x,t) - g(t)^2 \nabla_x \log p_t(x) \big) dt + g(t) d\bar w \label{eq:reverse_sde}
\end{align}
That is, SDE~\eqref{eq:reverse_sde} tells us how to run any SDE of the form~\eqref{eq:general_sde} backward in time! This means that we don't have to re-derive the reversal in each case, and we can choose any SDE solver to yield a practical sampler. But nothing is free: we sill cannot use~\eqref{eq:reverse_sde} directly to sample backward, since the term $\nabla_x \log p_t(x)$ -- which is in fact the \emph{score} that previously appeared in equation \ref{eqn:score} -- is unknown in general, since it depends on $p_t$. However, if we can \emph{learn} the score, then we can solve the reverse SDE. This is analogous to discrete diffusion, where the \emph{forward} process is easy to model (it just adds noise), while the \emph{reverse} process must be learned.

Let us take a moment to discuss the score, $\nabla_x \log p_t(x)$, which plays a central role. Intuitively, since the score ``points toward higher probability'', it helps to reverse the diffusion process, which ``flattens out'' the probability as it runs forward. The score is also related to the conditional expectation of $x_0$ given $x_t$. Recall that in the discrete case $$\sigma_q^2 \dt \nabla \log p_t(x_{t}) = \E[ x_{t-\dt} - x_t \mid x_{t} ] = \frac{\dt}{t} \E[ x_0 - x_t \mid x_{t} ],$$
(by equations \ref{eqn:score}, \ref{eq:var_reduc_dt}).

\newpage
Similarly, in the continuous case we have
\footnote{We can see this directly by applying Tweedie's formula, which states:
$$ \E[\mu_z | z] = z + \sigma_z^2 \nabla \log p(z) \text{ for } z \sim \cN(\mu_z, \sigma_z^2).$$
Since $x_t \sim \cN(x_0, t \sigma_q^2),$ Tweedie with $z \equiv x_t,$ $\mu_z \equiv x_0$ gives:
$$\E[x_0 | x_t] = x_t + t \sigma_q^2 \nabla \log p(x_t).$$
}
\begin{align}
    \sigma_q^2 \nabla \log p_t(x_{t}) 
    &= \frac{1}{t} \E[ x_0 - x_t \mid x_{t} ]. \label{eq:score_vs_expt_contin}
\end{align}

Returning to the reverse SDE, we can show that its discretization yields the DDPM sampler of Claim \ref{claim:ddpm_main} as a special case. The reversal of the simple SDE (\ref{eq:simple_sde}) is:
\begin{align}
    dx &= -\sigma_q^2 \nabla_x \log p_t(x) dt + \sigma_q d\bar w \\
    &= -\frac{1}{t} \E[ x_0 - x_t \mid x_{t} ] dt + \sigma_q d\bar w
\end{align}
The discretization is
\begin{align}
    x_t - x_{t-\dt} 
    &= -\frac{\dt}{t} \E[ x_0 - x_t \mid x_{t} ] + \cN(0, \sigma_q^2 \dt) \\
    &= - \E[ x_{t-\dt} - x_t \mid x_{t} ] + \cN(0, \sigma_q^2 \dt) \tag{by Eqn. \ref{eq:var_reduc_dt}} \\
    \implies
    x_{t-\dt}
    &= \E[ x_{t-\dt} \mid x_{t} ] + \cN(0, \sigma_q^2 \dt)
\end{align}
which is exactly the stochastic (DDPM) sampler derived in Claim \ref{claim:ddpm_main}.

\clearpage
\newpage
\section{Deterministic Sampling: DDIM}
\label{sec:ddim}

\newcommand{\rG}{{G_t}[x_0]}
\newcommand{\va}{v_t^{[a]}}
\newcommand{\vb}{v_t^{[b]}}
\newcommand{\vxn}{v_t^{[x_0]}}

We will now show a \emph{deterministic}
reverse sampler for Gaussian diffusion---
which appears similar to the stochastic sampler
of the previous section, but is conceptually quite different.
This sampler is equivalent to the DDIM\footnote{DDIM
stands for Denoising Diffusion Implicit Models,
which reflects a perspective used in the original derivation of \citet{song2021denoising}.
Our derivation follows a different perspective, 
and the ``implicit'' aspect will not be important to us.
}
update of \citet{song2021denoising},
adapted to in our simplified setting.

We consider the same Gaussian diffusion setup as the previous section,
with the joint distribution $(x_0, x_\dt, \dots, x_1)$
and conditional expectation function
$\mu_{t}(z) := \E[x_{t} \mid x_{t+\dt} = z].$
The reverse sampler is defined below, and listed explicitly in Pseudocode~\ref{alg:ddim_inf}.

\begin{mdframed}[nobreak=true]
\underline{Algorithm 2: Deterministic Reverse Sampler (DDIM-like)}\\
For input sample $x_t$, and step index $t$, output:
\begin{align}
\label{eqn:ddim}
    \hat{x}_{t-\dt} \gets
    x_t + \lambda({\mu}_{t-\dt}(x_t) - x_t)
\end{align}
where $\lambda := \left( \frac{\sigma_{t}}{\sigma_{t-\dt}+ \sigma_{t}} \right )$
and $\sigma_t \equiv \sigma_q \sqrt{t}$ 
from Equation~\eqref{eqn:intro-xt-dt}.
\end{mdframed}

How do we show that this defines a valid reverse sampler?
Since Algorithm 2 is \emph{deterministic},
it does not make sense to argue that it \emph{samples}
from $p(x_{t-\dt} \mid x_t)$, as we argued for the DDPM-like stochastic sampler.
Instead, we will directly show that Equation~\eqref{eqn:ddim}
implements a valid \emph{transport map}
between the marginal distributions $p_{t}$ and $p_{t-\dt}$.
That is, if we let $F_t$ be the update of Equation~\eqref{eqn:ddim}:
\begin{align}
    F_t(z) :=& ~z + \lambda(\mu_{t-\dt}(z) - z)\\
    =& ~z + \lambda(\E[x_{t-\dt} \mid x_t = z] - z)
\label{eqn:ddim_F}
\end{align}
then we want to show that\footnote{
The notation $F\sharp p$ means the
distribution of $\{F(x)\}_{x \sim p}$.
This is called the \emph{pushforward}
of $p$ by the function $F$.}
\begin{align}
F_t \sharp p_t \approx p_{t-\dt}.
\end{align}

{\bf Proof overview:}
The usual way to prove this
is to use tools from stochastic calculus, but 
we'll present an elementary derivation.
Our strategy will be to first show that Algorithm 2 is correct
in the simplest case of a point-mass distribution,
and then lift this result to full distributions
by marginalizing appropriately.
For the experts, this is similar to ``flow-matching'' proofs.

\subsection{Case 1: Single Point}
\label{sec:single_pt_ddim}

Let's first understand the simple case where 
the target distribution $p_0$ is a single point mass in $\R^d$.
Without loss of generality\footnote{Because we can just ``shift'' our coordinates to make it so. Formally, our entire setup including Equation~\ref{eqn:ddim_F} is translation-symmetric.},
we can assume the point is at $x_0 = 0$.
Is Algorithm 2 correct in this case?
To reason about correctness, we want to consider
the distributions of $x_t$ and $x_{t-\dt}$ for arbitrary step $t$.
According to the diffusion forward process (Equation~\ref{eqn:intro-fwd-dt}),
at time $t$ the relevant random variables are\footnote{We omit the Identity matrix in these covariances for notational simplicity. The reader may assume dimension $d=1$ without loss of generality.}
\begin{align*}
    x_0 &= 0 \quad\textrm{(deterministically)} \\
    x_{t-\dt} &\sim  \cN(x_0, \sigma_{t-\dt}^2) \\
    x_t &\sim  \cN(x_{t-\dt}, \sigma_{t}^2-\sigma_{t-\dt}^2).
\end{align*}
The marginal distribution of $x_{t-\dt}$ is $p_{t-\dt} = \cN(0, \sigma_{t-1}^2)$,
and the marginal distribution of $x_t$ is $p_t = \cN(0, \sigma_t^2)$.

Let us first find \emph{some} deterministic function $G_t: \R^d \to \R^d$,
such that
$G_t \sharp p_t = p_{t-\dt}$.
There are many possible functions which will work\footnote{For example,
we can always add a rotation around the origin to any valid map.}, but 
this is the obvious one:
\begin{align}
\label{eqn:gg}
    G_t(z) &:= \left(\frac{\sigma_{t-\dt}}{\sigma_t}\right) z.
\end{align}

The function $G_t$ above simply re-scales the Gaussian distribution of $p_t$,
to match variance of the Gaussian distribution $p_{t-\dt}$.
It turns out this $G_t$ is exactly equivalent to the
step $F_t$ taken by Algorithm 2,
which we will now show.

\begin{claim}
\label{claim:ddim_one_pt}
    When the target distribution is a point mass $p_0 = \delta_0$,
    then update $F_t$ (as defined in Equation~\ref{eqn:ddim_F})
    is equivalent to the scaling $G_t$ (as defined in Equation~\ref{eqn:gg}):
    \[
    F_t \equiv G_t.
    \]
Thus Algorithm 2 defines a reverse sampler for target distribution
$p_0 = \delta_{0}.$
\end{claim}

\begin{proof}
To apply $F_t$, we need to compute
$\E[x_{t-\dt} \mid x_t]$ for our simple distribution.
Since $(x_{t-\dt}, x_t)$ are jointly Gaussian, this is\footnote[][-2cm]{Recall
the conditional expectation of two jointly Gaussian random variables $(X, Y)$ is 
$\E[X \mid Y = y] = \mu_X + \Sigma_{XY} \Sigma^{-1}_{YY}(y-\mu_Y)$,
where $\mu_X, \mu_Y$ are the respective means,
and $\Sigma_{XY}, \Sigma_{YY}$ the cross-covariance of $(X, Y)$ and
covariance of $Y$.
Since $X = x_{t-\dt}$ and $Y = x_t$ are centered at $0$,
we have $\mu_X=\mu_Y=0$.
For the covariance term,
since $x_t = x_{t-\dt} + \eta$ we have
$\Sigma_{XY} = \E[x_t x_{t-\dt}^{T}] = \E[x_{t-\dt}x_{t-\dt}^{T}]
= \sigma_{t-\dt}^2 I_d$.
Similarly, $\Sigma_{YY} = \E[x_tx_t^T] = \sigma_t^2 I_d$.
}
\begin{align}
\label{line:exp}
    \E[x_{t-\dt} \mid x_t]
    &= \left(\frac{\sigma_{t-\dt}^2}{\sigma_t^2}\right) x_t.
\end{align}

The rest is algebra:
\begin{align*}
    F_t(x_t) &:= x_t + \lambda(\E[x_{t-\dt} \mid x_t ] - x_t)  \mathnote{by definition of $F_t$}\\
    &= 
    x_t + \left( \frac{\sigma_{t}}{\sigma_{t-\dt}+ \sigma_{t}}  \right)
    (\E[x_{t-\dt} \mid x_t ] - x_t) \mathnote{by definition of $\lambda$}\\
    &= 
    x_t + \left( \frac{\sigma_{t}}{\sigma_{t-\dt}+ \sigma_{t}}  \right)
    \left(\frac{\sigma_{t-\dt}^2}{\sigma_t^2} - 1\right)x_t \mathnote{by Equation~\eqref{line:exp}}\\
    &= \left(\frac{\sigma_{t-\dt}}{\sigma_t}\right) x_t  \\
    &= G_t(x_t).
\end{align*}
We therefore conclude that Algorithm 2
is a correct reverse sampler,
since it is equivalent to $G_t$, and $G_t$ is valid.
\end{proof}

The correctness of Algorithm 2 still holds\footnote{
See Claim~\ref{claim:ddim_x0} in Appendix~\ref{app:ddim} for an explicit statement.
}
if $x_0$ is an arbitrary point instead of $x_0 = 0$,
since everything is transitionally symmetric.

\subsection{Velocity Fields and Gases}
\label{sec:ddim-vf}

Before we move on, it will be helpful to think of
the DDIM update
as equivalent to a 
velocity field,
which moves points at time $t$ to their positions at time $(t-\dt)$.
Specifically, define the vector field
\[
\label{eqn:vt_ddim}
v_t(x_t) := \frac{\lambda}{\dt}(\E[x_{t-\dt} \mid x_t] - x_t).
\]
Then the DDIM update algorithm of Equation~\eqref{eqn:ddim}
can be written as:
\begin{align}
    \hat{x}_{t-\dt} :=& ~x_t + \lambda({\mu}_{t-\dt}(x_t) - x_t)
    \mathnote{from Equation~\eqref{eqn:ddim}}\\
        =& ~x_t + v_t(x_t) \dt.
\end{align}

The physical intuition for $v_t$ is: imagine a gas of non-interacting particles,
with density field given by $p_t$.
Then, suppose a particle at position $z$ moves in the direction $v_t(z)$.
The resulting gas will have density field $p_{t-\dt}$.
We write this process as
\[
p_t \goto{v_t} p_{t-\dt}.
\]
In the limit of small stepsize $\dt$, speaking informally,
we can think of $v_t$ as a \emph{velocity field} ---
which specifies the instantaneous velocity
of particles moving according to the DDIM algorithm.

\begin{marginfigure}
  \includegraphics[width=\linewidth]{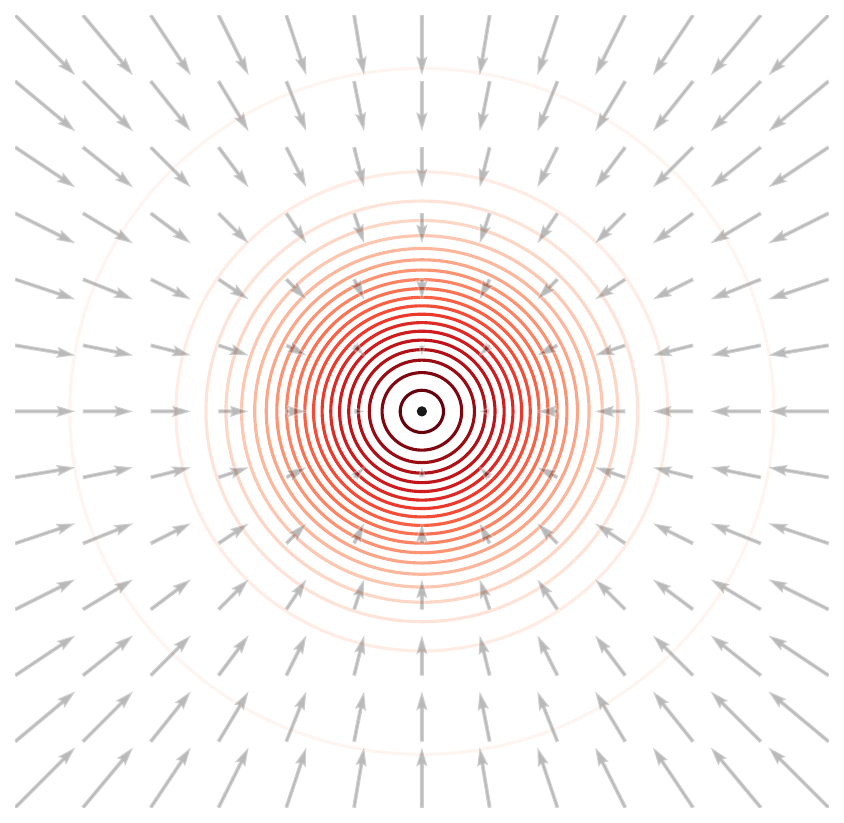}
  \caption{Velocity field $v_t$ when $p_0 = \delta_{x_0}$, overlaid
  on the Gaussian distribution $p_t$.}
  \label{fig:flow_single_pt}
\end{marginfigure}
As a concrete example, if the target distribution $p_0=\delta_{x_0}$, 
as in Section~\ref{sec:single_pt_ddim}, then the velocity field of DDIM is
$v_t(x_t) = \left( \frac{\sigma_{t} - \sigma_{t-\dt}}{\sigma_{t}} \right) (x_0 - x_t) / \dt$
which is a vector field that always points towards the initial point $x_0$
(see Figure~\ref{fig:flow_single_pt}).

\subsection{Case 2: Two Points}
  
Now let us show Algorithm 2 is correct when the target distribution is a mixture of two points:
\begin{equation}
\label{eqn:mixture}
p_0 := \frac{1}{2}\delta_a + \frac{1}{2}\delta_b,
\end{equation}
for some $a, b \in \R^d$.
According to the diffusion forward process,
the distribution at time $t$ will be a mixture of Gaussians\footnote{Linearity of the
forward process (with respect to $p_0$) was important here.
That is, roughly speaking, diffusing a distribution
is equivalent to diffusing each individual point in that distribution independently;
the points don't interact.
}:
\[
p_t := \frac{1}{2}\cN(a, \sigma_t^2) + \frac{1}{2}\cN(b, \sigma_t^2).
\]
We want to show that with these distributions $p_t$,
the DDIM velocity field $v_t$ (of Equation~\ref{eqn:vt_ddim})
transports
$p_t \goto{v_t} p_{t-\dt}$.

Let us first try to construct \emph{some}
velocity field $v^*_t$ such that
\mbox{$p_t \goto{v^*_t} p_{t-\dt}$}.
From our result in Section~\ref{sec:single_pt_ddim} --- the fact that DDIM update works for single points ---
we already know velocity fields which transport each mixture component $\{a, b\}$ individually.
That is, we know the
velocity field $\va$ defined as
\[
\label{eqn:def_va}
\va(x_t) := \lambda\E_{x_0 \sim \delta_a}[x_{t-\dt} - x_t \mid x_t]
\]
transports\footnote{Pay careful attention to which distributions
we take expectations over! The expectation in Equation \eqref{eqn:def_va}
is w.r.t. the single-point distribution $\delta_a$, 
but our definition of the DDIM algorithm,
and its vector field in Equation \eqref{eqn:vt_ddim},
are always w.r.t. the target distribution.
In our case, the target distribution is
$p_0$ of Equation~\eqref{eqn:mixture}.}
\[
\cN(a, \sigma_t^2)
\goto{\va}
\cN(a, \sigma_{t-\dt}^2),
\]
and similarly for $\vb$.

We now want some way of combining these two velocity fields 
into a single velocity $v^*_t$,
which transports the mixture:
\[
\label{eqn:guessT}
\underbrace{\left(\frac{1}{2}\cN(a, \sigma_t^2) + \frac{1}{2}\cN(b, \sigma_t^2) \right)}_{p_t}
\overset{v^*_t}{\longrightarrow} 
\underbrace{
\left(
\frac{1}{2}\cN(a, \sigma_{t-\dt}^2) + \frac{1}{2}\cN(b, \sigma_{t-\dt}^2)
\right)}_{p_{t-\dt}}
\]
We may be tempted to just take the average velocity field
$(v^*_t = 0.5\va + 0.5\vb)$, but this is incorrect.
The correct combined velocity $v_t^*$ 
is a \emph{weighted}-average of the individual velocity fields, 
weighted by their corresponding density fields\footnote{Note that we can write the density $\cN(x_t; a, \sigma_t^2)$ as $p(x_t \mid x_0=a)$.}.
\begin{align}
v_t^*(x_t)
&= \frac{\va(x_t) \cdot p(x_t \mid x_0 = a) + \vb(x_t) \cdot p(x_t \mid x_0 = b)}{ p(x_t \mid x_0 = a)  +  p(x_t \mid x_0 = b) } \\
&= \va(x_t) \cdot p(x_0 = a \mid x_t) + \vb(x_t) \cdot p(x_0 = b \mid x_t).
\label{eqn:weighting}
\end{align}
Explicitly, 
the weight for $\va$ at a point $x_t$
is the probability that $x_t$ was generated
from initial point $x_0=a$, rather than $x_0=b$.

\begin{figure*}[t]
    \classiccaptionstyle
    \centering
    \includegraphics[width=0.45\linewidth]{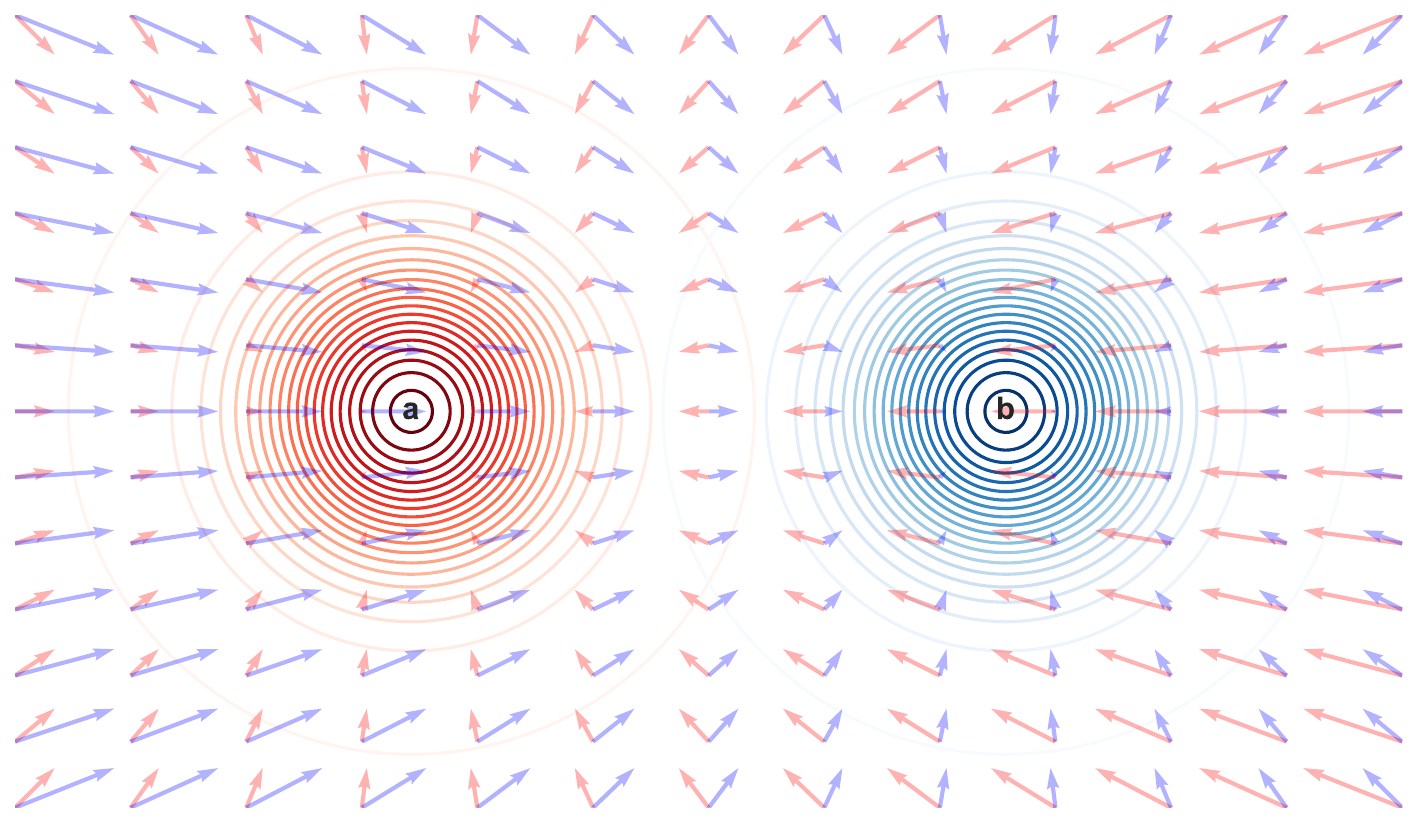}
    \hfill
    \includegraphics[width=0.45\linewidth]{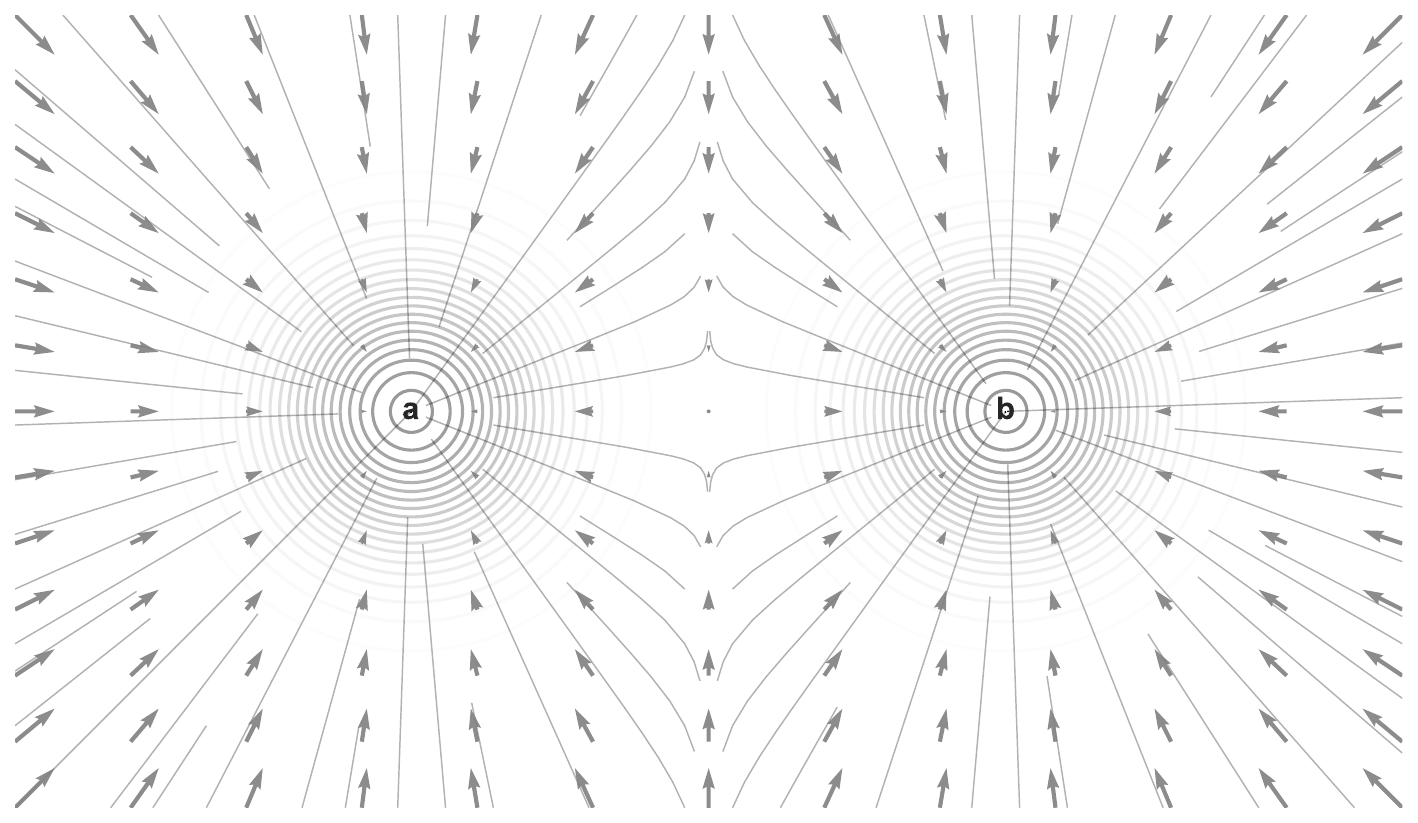}
    \caption{Illustration of combining the velocity fields of two gasses.
    Left: The density and velocity fields of two independent gases (in red and blue).
    Right: The effective density and velocity field of the combined gas,
    including streamlines.}
    \label{fig:gas}
\end{figure*}

To be intuitively convinced of this\footnote{The time step must be small enough 
for this analogy to hold, so the DDIM updates are essentially infinitesimal steps.
Otherwise, if the step size is large, it may not be possible to combine the two transport
maps with ``local'' (i.e. pointwise) operations alone.},
consider the corresponding question about gasses illustrated in Figure~\ref{fig:gas}.
Suppose we have two overlapping gases,
a red gas with density $\cN(a, \sigma^2)$ and velocity $\va$,
and a blue gas with density $\cN(b, \sigma^2)$ and
velocity $\vb$.
We want to know, what is the
effective velocity of the combined gas
(as if we saw only in grayscale)?
We should clearly take a \emph{weighted}-average of the individual gas velocities,
weighted by their respective densities --- just as in Equation~\eqref{eqn:weighting}.

We have now solved the main subproblem of this section: we have 
found \emph{one particular}
vector field $v_t^*$ which transports $p_t$ to $p_{t-\dt}$,
for our two-point distribution $p_0$.
It remains to show that this $v_t^*$ is equivalent
to the velocity field of Algorithm 2 ($v_t$ from Equation~\ref{eqn:vt_ddim}).

\newpage
To show this, first notice that the
individual vector field $\va$
can be written as a conditional expectation.
Using the definition in Equation~\eqref{eqn:def_va}\footnote{We add conditioning $x_0 = a$,
because we want to take expectations w.r.t
the two-point mixture distribution,
not the single-point distribution.},
\begin{align}
    \va(x_t) &= \lambda \E_{x_0 \sim \delta_a}[ x_{t-\dt} - x_t \mid x_{t} ] \\
    &= \lambda \E_{x_0 \sim \half\delta_a + \half\delta_b} [ x_{t-\dt} - x_t \mid x_0 = a, x_{t} ].
\end{align}
Now the entire vector field $v_t^*$
can be written as a conditional expectation:
\begin{align}
v_t^*(x_t)
&=
\va(x_t) \cdot p(x_0 = a \mid x_t)
+
\vb(x_t) \cdot p(x_0 = b \mid x_t) \\
&=
\lambda \E[ x_{t-\dt} - x_t \mid x_0 = a, x_{t}  ] \cdot p(x_0 = a \mid x_t ) \\
&\quad+ \lambda \E[ x_{t-\dt} - x_t \mid x_0 = b, x_{t}  ]  \cdot p(x_0 = b \mid x_t ) \\
&=
\lambda \E\left[
x_{t-\dt} - x_t \mid x_{t} \right] \\
&= v_t(x_t) \tag{from Equation~\ref{eqn:vt_ddim}}
\end{align}
where all expectations are w.r.t. the distribution $x_0 \sim \half\delta_a + \half\delta_b$.
Thus, the combined velocity field $v^*_t$ is exactly the
velocity field $v_t$ given by the updates of Algorithm 2 ---
so Algorithm 2 is a correct reverse sampler for our two-point mixture distribution.

\subsection{Case 3: Arbitrary Distributions}

Now that we know how to handle two points,
we can generalize this idea to
arbitrary distributions of $x_0$.
We will not go into details here, because the general proof will be subsumed
by the subsequent section.

It turns out that our overall proof strategy for Algorithm 2 can be generalized
significantly to other types of diffusions, without much work.
This yields the idea of flow matching, which we will see in the following section. Once we develop the machinery of flows, it is actually straightforward to derive DDIM directly from the simple single-point scaling algorithm of Equation~\eqref{eqn:gg}: see Appendix \ref{app:ddim_pf_by_flow}.

\subsubsection{The Probability Flow ODE [Optional]}
\label{sec:pf_ode}
Finally, we generalize our discrete-time deterministic sampler to an ordinary differential equation (ODE) called the \emph{probability flow ODE} \citep{song2020score}. The following section builds on our discussion of SDEs as the continuous limit of diffusion in section \ref{sec:sdes}. Just as the reverse-time SDEs of section \ref{sec:sdes} offered a flexible continuous-time generalization of discrete stochastic samplers, so we will see that discrete deterministic samplers generalize to ODEs. The ODE formulation offers both a useful theoretical lens through which to view diffusion, as well as practical advantages, like the opportunity to choose from a variety of off-the-shelf and custom ODE solvers to improve sampling (like the popular DPM++ method, as discussed in chapter \ref{sec:practical}).

Recall the general SDE (\ref{eq:general_sde}) from section \ref{sec:sdes}:
$$dx = f(x,t)dt + g(t)dw.$$
\citet{song2020score} showed that is possible to convert this SDE into a \emph{deterministic} equivalent called the probability flow ODE (PF-ODE):
\footnote{A proof sketch is in appendix \ref{append:sde}. It involves rewriting the SDE noise term as the deterministic score (recall the connection between noise and score in equation~\eqref{eqn:score}). Although it is deterministic, the score is unknown since it depends on $p_t$.}
\begin{align}
    \frac{dx}{dt} &= \tilde f(x, t), \quad \text{where } \tilde f(x, t) = f(x,t) - \frac{1}{2} g(t)^2 \nabla_x \log p_t(x) \label{eqn:pf_ode}
\end{align}
SDE~\eqref{eq:general_sde} and ODE~\eqref{eqn:pf_ode} are equivalent in the sense that trajectories obtained by solving the PF-ODE have the same marginal distributions as the SDE trajectories at every point in time\footnote{To use a gas analogy:
the SDE describes the (Brownian) motion of individual particles in a gas,
while the PF-ODE describes the streamlines of the gas's velocity field.
That is, the PF-ODE describes the motion of a ``test particle'' 
being transported by the gas--- like a feather in the wind.}.
However, note that the score appears here again, as it did in the reverse SDE \eqref{eq:reverse_sde}; just as for the reverse SDE, we must learn the score to make the ODE~\eqref{eqn:pf_ode} practically useful.

Just as DDPM was a (discretized) special-case of the reverse-time SDE~\eqref{eq:reverse_sde}, so DDIM can be seen as a (discretized) special case of the PF-ODE~\eqref{eqn:pf_ode}. Recall from section \ref{sec:sdes} that the simple diffusion we have been studying corresponds to the SDE (\ref{eq:simple_sde}) with $f = 0$ and $g = \sigma_q$. The corresponding ODE is
\begin{align}
    \frac{dx}{dt} &= -\frac{1}{2} \sigma_q^2 \nabla_x \log p_t(x) \\
    &= -\frac{1}{2t} \E[ x_0 - x_t \mid x_{t} ] \quad \text{(by eq. \ref{eq:score_vs_expt_contin})}
    \label{eq:simple_pf_ode} 
\end{align}
Reversing and discretizing yields
\begin{align*}
    x_{t-\dt} %
    &= x_{t} + \frac{\dt}{2t} \E[ x_0 - x_t \mid x_{t} ] \\
    &= x_{t} + \frac{1}{2} (\E[ x_{t-\dt} \mid x_{t} ] - x_{t}) \quad \text{(by eq. \ref{eq:var_reduc_dt}).}
\end{align*}
Noting that $\lim_{\dt \to 0} \left( \frac{\sigma_{t}}{\sigma_{t-\dt}+ \sigma_{t}} \right ) = \frac{1}{2}$, we recover the deterministic (DDIM) sampler (\ref{eqn:ddim}).

\subsection{Discussion: DDPM vs DDIM}

The two reverse samplers defined above (DDPM and DDIM)
are conceptually significantly different:
one is deterministic, and the other stochastic.
To review, these samplers use the following strategies:
\begin{enumerate}
    \item DDPM ideally implements a stochastic map $F_t$,
    such that the output $F_t(x_{t})$ is, pointwise,
    a sample from the conditional distribution
    $p(x_{t-\dt} \mid x_{t})$.
    \item DDIM ideally implements a deterministic map $F_t$,
    such that the output $F_t(x_{t})$
    is \emph{marginally distributed} as $p_{t-\dt}$.
    That is, \mbox{$F_t \sharp p_{t} = p_{t-\dt}$}.
\end{enumerate}

Although they both happen to take steps in the same direction\footnote{Steps proportional to $({\mu}_{t-\dt}(x_t) - x_t)$.}
(given the same input $x_t$),
the two algorithms end up evolving very differently.
To see this, let's consider how each sampler ideally behaves, when started
from the same initial point $x_1$ and iterated to completion.

DDPM will ideally produce a sample from $p(x_0 \mid x_1)$.
If the forward process mixes sufficiently (i.e. for large $\sigma_q$ in our setup),
then the final point $x_1$ will be nearly \emph{independent} from the initial point.
Thus $p(x_0 \mid x_1) \approx p(x_0)$, so the distribution output by 
the ideal DDPM will not depend at all\footnote{
Actual DDPMs may have a small dependency on the initial point $x_1$,
because they do not mix perfectly
(i.e. the final distribution $p_1$ is not perfectly Gaussian).
Randomizing the initial point may thus help with sample diversity in practice.} on the starting point $x_1$.
In contrast, DDIM is deterministic, so it will always produce a 
fixed value for a given $x_1$, and thus will depend very strongly on $x_1$.

The picture to have in mind is, DDIM defines a deterministic map $\R^d \to \R^d$,
taking samples from a Gaussian distribution to our target distribution.
At this level, the DDIM map may sound similar to other generative models --- after all,
GANs and Normalizing Flows also define maps from Gaussian noise to the true distribution.
What is special about the DDIM map is, it is not allowed to be arbitrary:
the target distribution $p^*$ exactly determines the ideal DDIM map
(which we train models to emulate).
This map is ``nice''; for example we expect it to be smooth if our target distribution is smooth.
GANs, in contrast, are free to learn any arbitrary mapping between noise and images.
This feature of diffusion models may make the learning problem easier
in some cases (since it is supervised), or harder in other cases
(since there may be easier-to-learn maps which other methods could find).

\subsection{Remarks on Generalization}
\label{sec:generalization}

In this tutorial, we have not discussed the learning-theoretic 
aspects of diffusion models:
How do we learn properties of the underlying distribution,
given only finite samples and bounded compute?
These are fundamental aspects of learning,
but are not yet fully understood for diffusion models;
it is an active area of research\footnote{
We recommend the introductions of \citet{chen2022sampling} 
and \citet{chen2024learning} for an overview of
recent learning-theoretic results.
This line of work includes e.g.
\citet{de2021diffusion,de2022convergence,lee2023convergence,chen2023improved,chen2024probability}.
}.

To appreciate the subtlety here, suppose we learn a diffusion model
using the classic strategy
of Empirical Risk Minimization (ERM): 
we sample a finite train set from the underlying distribution,
and optimize all regression functions w.r.t. this empirical
distribution.
The problem is, we should not \emph{perfectly}
minimize the empirical risk, because this would yield a diffusion model
which only reproduces the train samples\footnote{
This is not specific to diffusion models:
any perfect generative model of the empirical distribution
will always output a uniformly random train point,
which is far-from-optimal w.r.t. the true underlying distribution.}.

In general learning the diffusion model must be \emph{regularized},
implicitly or explicitly, to prevent overfitting and memorization of the training data.
When we train deep neural networks for use in diffusion models,
this regularization often occurs implicitly:
factors such as finite model size and optimization randomness prevent the trained model
from perfectly memorizing its train set.
We will revisit these factors (as sources of error)
in Section~\ref{sec:practical}.

This issue of memorizing training data has been seen ``in the wild'' in diffusion models
trained on small image datasets, and it has been observed that memorization reduces as the training set size increases \citep{somepalli2023diffusion,gu2023memorization}.  Additionally, memorization as been noted as a potential security and copyright issue for neural networks as in \citet{carlini2023extracting} where the authors found they can recover training data from stable diffusion with the right prompts.

\begin{figure*}[p]
\classiccaptionstyle
    \centering
    \includegraphics[width=0.95\linewidth]{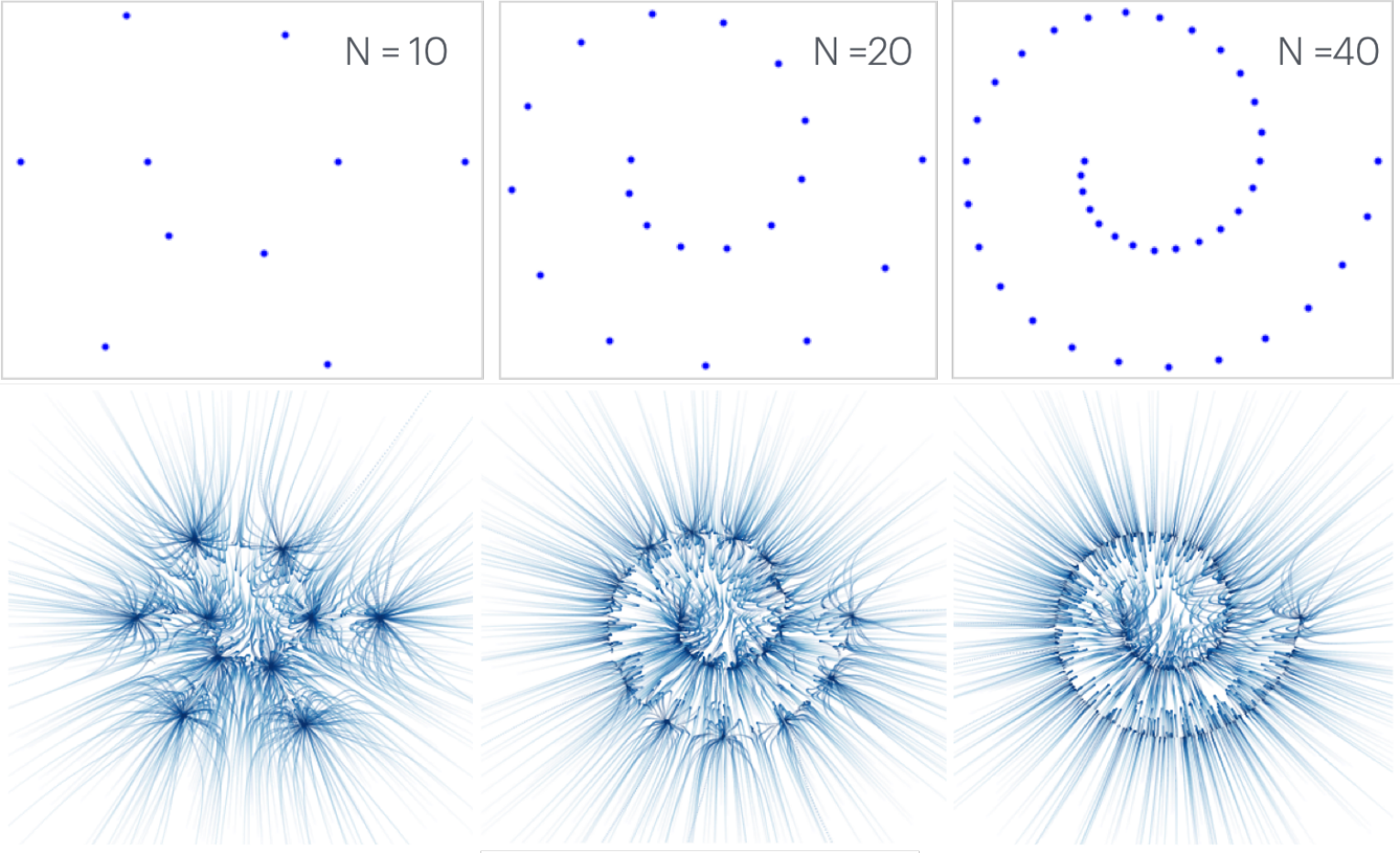}
    
    \caption{The DDIM trajectories (shaded by timestep $t$) for a spiral dataset. We compare the trajectories with $10, 20$, and $40$ training samples. Note that as we add more training points (moving left to right) the diffusion algorithm begins to \textit{learn} the underlying spiral and the trajectories look more perpendicular to the underlying manifold. The network used here is a 3 layer ReLU network with 128 neurons per layer.}
    \label{fig:spiders}
\end{figure*}

Figure~\ref{fig:spiders} demonstrates the effect of training set size, and shows the DDIM trajectories
for a diffusion model trained using a 3 layer ReLU network.
We see that the diffusion model on $N=10$ samples ``memorizes'' its train set: its trajectories all collapse to one of the train points,
instead of producing the underlying spiral distribution.
As we add more samples,
the model starts to generalize:
the trajectories converge to the underlying spiral manifold.
The trajectories also start to become more
perpendicular the underlying manifold,
suggesting that the low dimensional structure is being learned.
We also note that in the $N=10$ case where the diffusion model fails,
it is not at all obvious a human would be able
to identify the ``correct'' pattern from these samples,
so generalization may be too much to expect.

\clearpage
\newpage
\section{Flow Matching}
\label{sec:flows}

\begin{marginfigure}
  \includegraphics[height=\textheight,keepaspectratio]{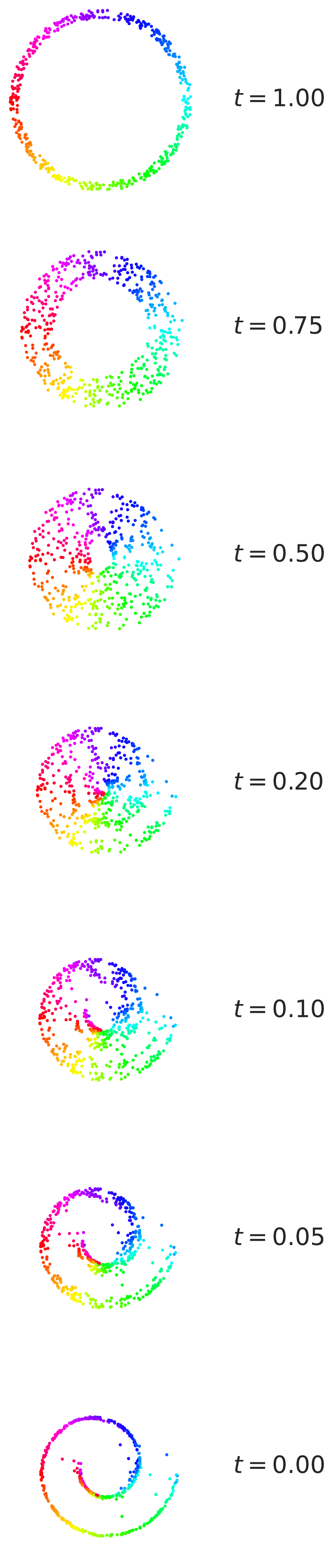}
  \caption{Running a flow which generates a spiral distribution (bottom) from
  an annular distribution (top).}
  \label{fig:cfm_demo}
\end{marginfigure}

We now introduce the framework of flow matching \citep{peluchetti2022nondenoising,liu2022let,liu2022flow,lipman2023flow,albergo2023stochastic}. %
Flow matching 
can be thought of as a generalization of DDIM,
which allows for more flexibility in designing
generative models--- including for example the \emph{rectified flows} 
(sometimes called  \emph{linear flows}) 
used by Stable Diffusion 3 \citep{liu2022flow,esser2024scaling}.

We have actually already seen the main ideas behind flow matching,
in our analysis of DDIM in Section~\ref{sec:ddim}.
At a high level, here is how we constructed a generative 
model in Section~\ref{sec:ddim}:
\begin{enumerate}
    \item First, we defined how to generate a single point.
    Specifically, we constructed vector fields $\{\va\}_t$ which,
    when applied for all time steps, transported
    a standard Gaussian distribution to an arbitrary delta distribution $\delta_a$.
    \item Second, we determined how to combine two vector fields
    into a single effective vector field. This lets us construct 
    a transport from the standard Gaussian to
    \emph{two} points (or, more generally,
    to a \emph{distribution} over points --- our target distribution).
\end{enumerate}
Neither of these steps particularly require the Gaussian base distribution,
or the Gaussian forward process (Equation~\ref{eqn:intro-fwd}).
The second step of combining vector fields 
remains identical for any two arbitrary vector fields, for example.

So let's drop all the Gaussian assumptions. Instead, we will begin by thinking at a basic level about how to map between any two points $x_0$ and $x_1$. Then, we see what happens when the two points are sampled from arbitrary distributions $p$ (data) and $q$ (base), respectively. We will see that this point of view encompasses DDIM as a special case, but that it is significantly more general.

\subsection{Flows}
Let us first define the central notion of a \emph{flow}.
A \emph{flow} is simply a collection
of time-indexed vector fields $v = \{v_t\}_{t \in [0, 1]}$.
We should think of this as the velocity-field $v_t$ of
a gas at each time $t$, as we did earlier in Section~\ref{sec:ddim-vf}.
Any flow defines a trajectory taking
initial points $x_1$
to final points $x_0$, 
by transporting the initial point along the velocity fields $\{v_t\}$.

\newpage
Formally, for flow $v$ and initial point $x_1$, consider the ODE\footnote[][]{
The corresponding discrete-time analog is the iteration:
\mbox{$x_{t - \dt} \gets x_t + v_t(x_t)\dt$},
starting at $t=1$ with initial point $x_1$.
}
\begin{align}
    \frac{dx_t}{dt} &= -v_t(x_t), \label{eqn:pt_ode}
\end{align}
with initial condition $x_1$ at time $t=1$.
We write
\begin{align}
x_t := \mathrm{RunFlow}(v, x_1, t)
\end{align}
to denote the solution to the flow ODE (Equation~\ref{eqn:pt_ode}) at time $t$,
terminating at final point $x_0$.
That is, RunFlow is the result of transporting point $x_1$ along the flow $v$
up to time $t$.

Just as flows define maps between initial and final points,
they also define transports between entire \emph{distributions},
by ``pushing forward'' points from the source distribution
along their trajectories.
If $p_1$ is a distribution on initial points\footnote{Notational warning: Most of the flow matching literature
uses a reversed time convention, so $t=1$ is the target distribution. We let $t=0$ be the target distribution
to be consistent with the DDPM convention.},
then applying the flow $v$ yields the distribution on final points\footnote{
We could equivalently write this as the pushforward $\rflow(v, \cdot, 0) \sharp p_1$.
}
\[
p_0 = \{\rflow(v, x_1, t=0)\}_{x_1 \sim p_1}.
\]
We denote this process as 
$
p_1 \flowto{v} p_0
$
meaning the flow $v$ transports initial distribution $p_1$
to final distribution\footnote{
In our gas analogy, this means if we start with a gas
of particles distributed according to $p_1$,
and each particle follows the trajectory defined by $v$,
then the final distribution of particles will be $p_0$.}
$p_0$.

\newthought{The ultimate goal of flow matching} is to somehow learn a flow $v^*$
which transports $q \flowto{v^*} p$, where $p$ is the target distribution
and $q$ is some easy-to-sample base distribution (such as a Gaussian).
If we had this $v^*$, we could generate samples from our target $p$
by first sampling $x_1 \sim q$, then running our flow 
with initial point $x_1$ and outputting the resulting final point $x_0$.
The DDIM algorithm of Section~\ref{sec:ddim} was actually a
special case\footnote{To connect to diffusion: The continuous-time limit of DDIM (\ref{eq:simple_pf_ode}) is a flow with $v_t(x_t) = \frac{1}{2t} \E[x_0 - x_t |x_t]$. The base distribution $p_1$ is Gaussian. DDIM Sampling (algorithm \ref{alg:ddim_inf}) is a discretized method for evaluating RunFlow. DDPM Training (algorithm \ref{alg:ddpm_gen}) is a method for learning $v^\star$ -- but it relies on the Gaussian structure and differs somewhat from the flow-matching algorithm we will present in this chapter.} of this, for a very particular choice of
flow $v^*$.
Now, how do we construct such flows in general?

\subsection{Pointwise Flows}
\label{sec:ptwise-flow}

\begin{marginfigure}
  \includegraphics[width=\textwidth,keepaspectratio]{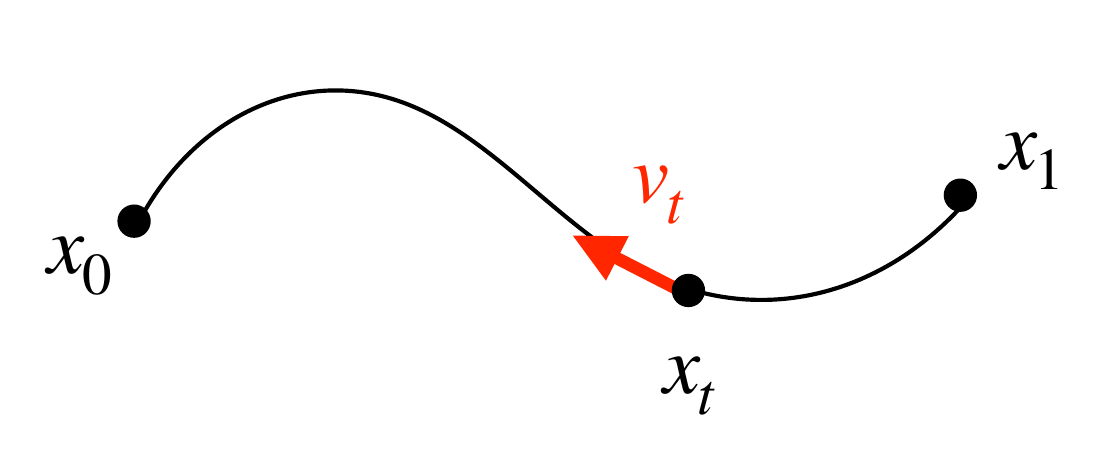}
  \caption{A pointwise flow $\Tab{x_1}{x_0}_t$ transporting $x_1$ to $x_0$.}
  \label{fig:pt_flow}
\end{marginfigure}

Our basic building-block will be a \emph{pointwise flow}
which just transports a single point $x_1$ to a point $x_0$. 
Intuitively, given an arbitrary path $\{x_t\}_{t \in [0, 1]}$
that connects $x_1$ to $x_0$,
a pointwise flow describes this trajectory
by giving its velocity $v_t(x_t)$ at each point $x_t$ along it
(see Figure~\ref{fig:pt_flow}).
Formally, a \emph{pointwise flow between $x_1$ and $x_0$}
is any flow $\{ v_t \}_t$ that satisfies Equation~\ref{eqn:pt_ode} 
with boundary conditions $x_1$ and $x_0$ at times $t=1, 0$ respectively.
We denote such flows as $\Tab{x_1}{x_0}$.
Pointwise flows are not unique:
there are many different choices of path between $x_0$ and $x_1$.

\subsection{Marginal Flows}
Suppose that for all pairs of points
$(x_1, x_0)$,
we can construct an explicit pointwise flow $\Tab{x_1}{x_0}$ that transports a
source point $x_1$ to target point $x_0$.
For example, we could let $x_t$ travel along a straight line from $x_1$ to $x_0$, or along any other explicit path.
Recall in our gas analogy, this corresponds to an individual particle
that moves between $x_1$ and $x_0$.
Now, let us try to set up a collection of individual particles, such that
at $t=1$ the particles are distributed according to $q$,
and at $t=0$ they are distributed according to $p$.
This is actually easy to do:
We can pick any coupling\footnote{A \emph{coupling} $\Pi_{q, p}$ between $q$ and $p$, specifies how to jointly sample pairs $(x_1, x_0)$ of source and target points, such that $x_0$ is marginally distributed as $p$, and $x_1$ as $q$. The most basic coupling is the \emph{independent coupling}, with corresponds to sampling $x_1, x_0$ independently.}
$\Pi_{q,p}$ between $q$ and $p$,
and consider particles corresponding to
the pointwise flows $\{ \Tab{x_1}{x_0} \}_{(x_1, x_0) \sim \Pi_{q,p}}$.
This gives us a distribution over pointwise flows
(i.e. a collection of particle trajectories) with the desired behavior in aggregate.

We would like to combine all of these pointwise flows somehow, to get
a single flow $v^*$ that implements the same transport between distributions\footnote{
Why would we like this? As we will see later, it simplifies our learning problem:
instead of having to learn the distribution of all the individual trajectories, we can instead just learn one velocity field representing their bulk evolution.}.
Our previous discussion\footnote{Compare to Equation~\eqref{eqn:weighting} in Section~\ref{sec:ddim}.
A formal statement of how to combine flows is given in Appendix~\ref{sec:flow_formal}.} in Section~\ref{sec:ddim}
tells us how to do this: to determine the effective velocity $v_t^*(x_t)$,
we should
take a weighted-average of all individual particle velocities $\Tab{x_1}{x_0}_t$,
weighted by the probability that a particle at $x_t$ 
was generated by the pointwise flow $\Tab{x_1}{x_0}$.
The final result is\footnote{
An alternate way of viewing this result at a high level is:
we start with pointwise flows $\Tab{x_1}{x_0}$
which transport delta distributions:
\[
\delta_{x_1} \flowto{\Tab{x_1}{x_0}} \delta_{x_0}.
\]
And then Equation~\eqref{eqn:vexp} gives us a fancy way of
``averaging these flows over $x_1$ and $x_0$'', 
to get a flow $v^*$ transporting
\[
q = \E_{x_1 \sim q}[\delta_{x_1}] \flowto{v^*} \E_{x_0 \sim p}[\delta_{x_0}] = p.
\]
}
\begin{align}
v^*_t(x_t) &:= \E_{x_0, x_1 \mid x_t}[\Tab{x_1}{x_0}_t(x_t) \mid x_t] 
\label{eqn:vexp}
\end{align}
where the expectation is w.r.t. the joint distribution of $(x_1, x_0, x_t)$
induced by sampling $(x_1, x_0) \sim \Pi_{q,p}$ and letting
$x_t \gets \mathrm{RunFlow}(\Tab{x_1}{x_0}, x_1, t)$.

At this point, we have a ``solution'' to our generative modeling problem in principle,
but some important questions remain to make it useful in practice:
\begin{itemize}
    \item Which pointwise flow
    $\Tab{x_1}{x_0}$ and coupling $\Pi_{q,p}$ should we chose?
    \item How do we compute the marginal flow $v^*$?
    We cannot compute it from
    Equation~\eqref{eqn:vexp} directly, because this would require
    sampling from $p(x_0 \mid x_t)$ for a given point $x_t$,
    which may be complicated in general.
\end{itemize}
We answer these in the next sections.

\subsection{A Simple Choice of Pointwise Flow}
\begin{figure*}[t]
    \centering
    \classiccaptionstyle
    \includegraphics[width=0.95\linewidth]{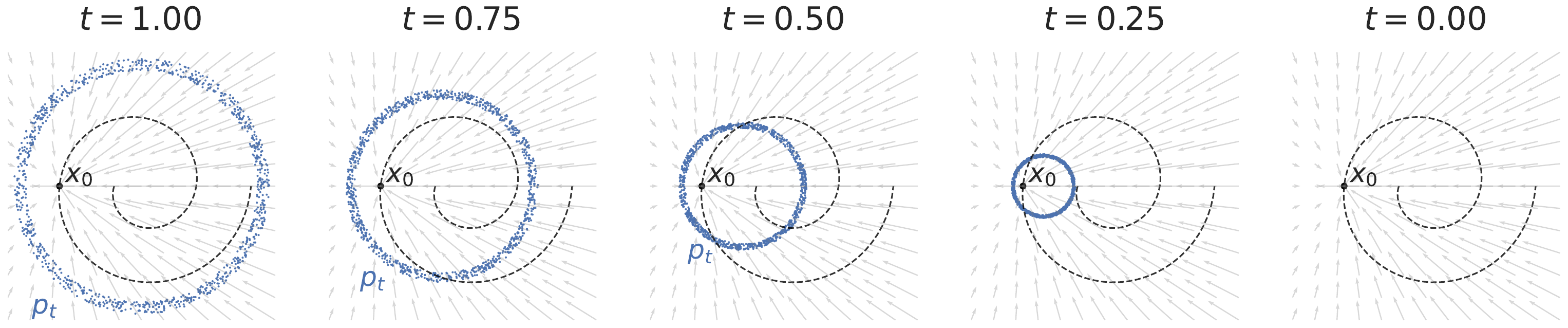}
    \caption{A marginal flow with linear pointwise flows, base distribution $q$ uniform over an annulus, and target distribution $p$ equal to a Dirac-delta at $x_0$. (This can also be thought of as the average over $x_1$ of the pointwise linear flows from \mbox{$x_1 \sim q$} to a fixed $x_0$).
    Gray arrows depict the flow field at different times $t$.
    The leftmost \mbox{$(t=1)$} plot shows samples from the base distribution $q$.
    Subsequent plots show these samples transported by the flow
    at intermediate times $t$, 
    The final \mbox{$(t=0)$} plot shows
    all points collapsed to the target $x_0$.
    This particular $x_0$ happens to be one point on the spiral distribution of Figure~\ref{fig:cfm_demo}.
    }
    \label{fig:circle-cfm}
\end{figure*}

We need an explicit choices of: pointwise flow,
base distribution $q$, and coupling $\Pi_{q, p}$.
There are many simple choices which would work\footnote{
Diffusion provides one possible construction, as we will see later in Section \ref{sec:ddim_flow}.}.

The base distribution $q$ can be essentially any easy-to-sample distribution.
Gaussians are a popular choice but certainly not the only one---
Figure \ref{fig:cfm_demo} uses an annular base distribution, for example.
As for the coupling $\Pi_{q,p}$
between the base and target distribution, the simplest choice
is the independent coupling, i.e. sampling from $p$ and $q$ independently.

For a pointwise flow, arguably the simplest construction is a \emph{linear pointwise flow}:
\begin{align}
    \Tab{x_1}{x_0}_t(x_t) &= x_0 - x_1, \label{ln:linear_ptwise}\\
    \implies \mathrm{RunFlow}(\Tab{x_1}{x_0}, x_1, t) &= tx_1 + (1-t)x_{0}
    \label{eqn:lin_runflow}
\end{align}
which simply linearly interpolates between $x_1$ and $x_0$ (and corresponds to the choice made in \citet{liu2022flow}).
In Figure \ref{fig:circle-cfm} we visualize a marginal flow composed of linear pointwise flows, the same annular base distribution $q$ of Figure \ref{fig:cfm_demo}, and target distribution equal to a point-mass ($p = \delta_{x_0}$)\footnote{A marginal distribution with a point-mass target distribution -- or equivalently the average of pointwise flows over the the base distribution only -- is sometimes called a \emph{(one-sided) conditional flow} \citep{lipman2023flow}. 
}.

\subsection{Flow Matching}
Now, the only remaining problem is that naively
evaluating $v^*$ using Equation~\eqref{eqn:vexp} requires sampling from $p(x_0 \mid x_t)$ for a given $x_t$. If we knew how do this for $t=1$, we would have already solved the generative modeling problem!

Fortunately, we can take advantage of the same trick
from DDPM: it is enough for us to be able to sample from the joint distribution $(x_0, x_t)$,
and then solve a regression problem.
Similar to DDPM, the conditional expectation function in Equation~\eqref{eqn:vexp}
can be written as a regressor\footnote{This result is analogous to Theorem 2 in \citet{lipman2023flow}, but ours is for a two-sided flow.}:

\begin{align}
v^*_t(x_t) &:= \E_{x_0, x_1 \mid x_t}[\Tab{x_1}{x_0}_t(x_t) \mid x_t] \\
\implies 
v_t^* 
&=\argmin_{f: \R^d \to \R^d} ~~\E_{(x_0, x_1, x_t)}
||f(x_t) -  \Tab{x_1}{x_0}_t(x_t)||_2^2,
\label{eqn:cfm_ptwise}
\end{align}
(by using the generic fact that $\argmin_{f} \E ||f(x) - y||^2 = \E[y \mid x]$).

\bigskip
In words, Equation~\eqref{eqn:cfm_ptwise}
says that to compute the loss of a model $f_\theta$ for a fixed time $t$, we should:
\begin{enumerate}
    \item Sample source and target points $(x_1, x_0)$ from their joint distribution.
    \item Compute the point $x_t$ deterministically, by running\footnote{
    If we chose linear pointwise flows, for example, this would mean
    \mbox{$x_t \gets tx_1 + (1-t)x_0$}, via Equation~\eqref{eqn:lin_runflow}.
    }
    the pointwise flow
    $\Tab{x_1}{x_0}$ starting from point $x_1$ up to time $t$.
    \item Evaluate the model's prediction at $x_t$, as $f_\theta(x_t)$.
    Evaluate the deterministic vector $\Tab{x_1}{x_0}_t(x_t)$. 
    Then compute L2 loss between these two quantities. 
\end{enumerate}

To sample from the trained model (our estimate of $v_t^*$),
we first sample a source point $x_1 \sim q$, then transport it along the learnt flow
to a target sample $x_0$. Pseudocode listings ~\ref{alg:fm_train} and ~\ref{alg:fm_gen}
give the explicit procedures for training and sampling from flow-based models
(including the special case of linear flows for concreteness; matching Algorithm 1 in \citet{liu2022flow}.).

\subsection*{Summary}
To summarize, here is how to learn a flow-matching generative model for target distribution $p$.

\paragraph{The Ingredients.}
We first choose:
\begin{enumerate}
    \item A source distribution $q$, from
    which we can efficiently sample
    (e.g. a standard Gaussian).
    \item A \emph{coupling} $\Pi_{q, p}$ between 
    $q$ and $p$, which specifies
    a way to jointly sample a pair of source and target points $(x_1, x_0)$
    with marginals $q$ and $p$ respectively.
    A standard choice is the independent coupling, i.e.
    sample $x_1 \sim q$ and $x_0 \sim p$ independently.
    
    \item For all pairs of points $(x_1, x_0)$,
    an explicit \emph{pointwise flow} $\Tab{x_1}{x_0}$
    which transports $x_1$ to $x_0$.
    We must be able to efficiently compute the vector field $\Tab{x_1}{x_0}_t$
at all points.
\end{enumerate}
These ingredients determine, in theory, a marginal vector field $v^*$
which transports $q$ to $p$:
\begin{align}
\label{eqn:vstar}
v^*_t(x_t) &:= \E_{x_0, x_1 \mid x_t}[\Tab{x_1}{x_0}_t(x_t) \mid x_t]
\end{align}
where the expectation is w.r.t. the joint distribution:
\begin{align*}
(x_1, x_0) &\sim \Pi_{q,p} \\
x_t &:= \mathrm{RunFlow}(\Tab{x_1}{x_0}, x_1, t).
\end{align*}

\paragraph{Training.}
Train a neural network $f_\theta$ by backpropogating the 
stochastic loss function computed by Pseudocode~\ref{alg:fm_train}.
The optimal function for this expected loss is:
$f_\theta(x_t, t) = v^*_t(x_t)$.

\paragraph{Sampling.}
Run Pseudocode~\ref{alg:fm_gen} to generate a sample $x_0$
from (approimately) the target distribution $p$.

\SetKwComment{Comment}{// }{}
\begin{fullwidth}
\begin{minipage}[t]{0.47\linewidth}
\begin{algorithm}[H]
\DontPrintSemicolon
\caption{Flow-matching train loss, generic pointwise flow \textcolor{red}{[or linear flow]} }\label{alg:fm_train}
\KwIn{Neural network $f_\theta$}
\KwData{Sample-access to coupling $\Pi_{q, p}$;
Pointwise flows $\{\Tab{x_1}{x_0}_t\}$ for all $x_1, x_0$.}
\KwOut{Stochastic loss $L$}
$(x_1, x_0) \gets \mathrm{Sample}( \Pi_{q,p} )$ \;
$t \gets \textrm{Unif}[0, 1]$ \;
$x_t \gets \mathcolor{red}{\underbrace{\mathcolor{black}{\mathrm{RunFlow}(\Tab{x_1}{x_0}, x_1, t)}}_{tx_1 + (1-t)x_{0}}}$  \;
$L \gets \Big\Vert f_\theta(x_t, t) -  \mathcolor{red}{\underbrace{\mathcolor{black}{\Tab{x_1}{x_0}_t(x_t)}}_{(x_0 - x_1)}} \Big\Vert_2^2$ \;
\Return $L$
\end{algorithm}
\end{minipage}
\hfill
\begin{minipage}[t]{0.47\linewidth}
\begin{algorithm}[H]
\DontPrintSemicolon
\caption{Flow-matching sampling}\label{alg:fm_gen}
\KwIn{Trained network $f_\theta$}
\KwData{Sample-access to base distribution $q$; step-size $\dt$.}
\KwOut{Sample from target distribution $p$.}
$x_1 \gets \mathrm{Sample}(q)$ \;
\For{$t=1,~(1-\dt),(1-2\dt), \dots, \dt$}{
$x_{t-\dt} \gets x_t + f_\theta(x_t, t)\dt$ \;
}
\Return $x_0$
\end{algorithm}
\end{minipage}

\end{fullwidth}

\subsection{DDIM as Flow Matching [Optional]}
\label{sec:ddim_flow}
The DDIM algorithm of Section~\ref{sec:ddim}
can be seen as a special case of flow matching,
for a particular choice of pointwise flows
and coupling.
We describe the exact correspondence here, which will allow us to notice an interesting
relation between DDIM and linear flows.

We claim DDIM is equivalent to flow-matching with the following parameters: 
\begin{enumerate}
    \item {\bf Pointwise Flows:} Either of the two equivalent pointwise flows:
    \[
    \label{eqn:ptwise_ddim}
    \Tab{x_1}{x_0}_t(x_t) := \frac{1}{2t} (x_t - x_0)
    \]
    or 
    \[
    \label{eqn:ptwise_ddim2}
    \Tab{x_1}{x_0}_t(x_t)
    := \frac{1}{2\sqrt{t}}(x_0 - x_1),
    \]
    which both generate the trajectory\footnote{
See Appendix~\ref{app:ddim_two_ptwise_flows_equiv} for details on why ~\eqref{eqn:ptwise_ddim} and~\eqref{eqn:ptwise_ddim2} are equivalent along their trajectories.}:
    \[
    \label{eqn:ptwise_traj}
    x_t = x_0 + (x_1 - x_0)\sqrt{t}.
    \]
    
    \item {\bf Coupling:}
    The ``diffusion coupling'' -- that is, the joint distribution 
    on $(x_0, x_1)$ generated by
    \[
    \label{eqn:ddpm_coupling}
    x_0 \sim p; \quad x_1 \gets x_0 + \cN(0, \sigma_q^2).
    \]
\end{enumerate}
This claim is straightforward to prove
(see Appendix~\ref{app:ddim_pf_by_flow}),
but the implication is somewhat surprising:
we can recover the DDIM trajectories
(which are not straight in general)
as a combination of the \emph{straight}
pointwise trajectories in Equation~\eqref{eqn:ptwise_traj}.
In fact,
the DDIM trajectories
are exactly equivalent to flow-matching trajectories
for the above linear flows, with a different scaling
of time ($\sqrt{t}$ vs. $t$)\footnote{
DDIM at time $t$ corresponds to the linear flow at time $\sqrt{t}$;
thus linear flows are ``slower'' than DDIM when $t$ is small.
This may be beneficial for linear flows in practice (speculatively).}.
\begin{mdframed}[nobreak]
\begin{claim}[DDIM as Linear Flow; Informal]
\label{claim:ddim-flow}
{The DDIM sampler (Algorithm 2)}
is equivalent, up to time-reparameterization,
to the marginal flow produced by
linear pointwise flows (Equation~\ref{ln:linear_ptwise}) with the diffusion coupling
(Equation~\ref{eqn:ddpm_coupling}).
\end{claim} 
\end{mdframed}
A formal statement of this claim\footnote{
In practice, linear flows are most often instantiated with
the independent coupling, not the above ``diffusion coupling.''
However, for large enough terminal variance $\sigma_q^2$,
the diffusion coupling is close to independent.
Therefore, Claim~\ref{claim:ddim-flow} tells us
that the common practice in flow matching
(linear flows with a Gaussian terminal distribution and independent coupling)
is nearly equivalent to standard DDIM, with a different time schedule.
Finally, for the experts:
this is a claim about the ``variance exploding'' version of DDIM,
which is what we use throughout. 
Claim~\ref{claim:ddim-flow} is false for variance-preserving DDIM.
}
is provided in Appendix~\ref{app:ddim_sqrt_time}.

\subsection{Additional Remarks and References [Optional]}

\begin{marginfigure}[1cm]
  \includegraphics[width=\textwidth,keepaspectratio]{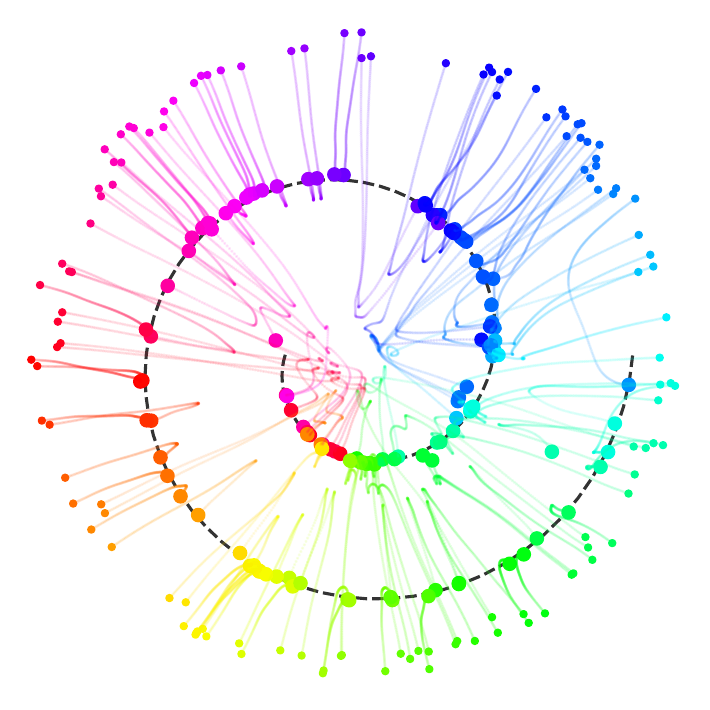}
  \caption{The trajectories of individual samples \mbox{$x_1 \sim q$} for the flow in Figure~\ref{fig:cfm_demo}.}
  \label{fig:cfm_traj}
\end{marginfigure}

\begin{figure*}
    \classiccaptionstyle
    \centering
    \includegraphics[width=0.98\linewidth]{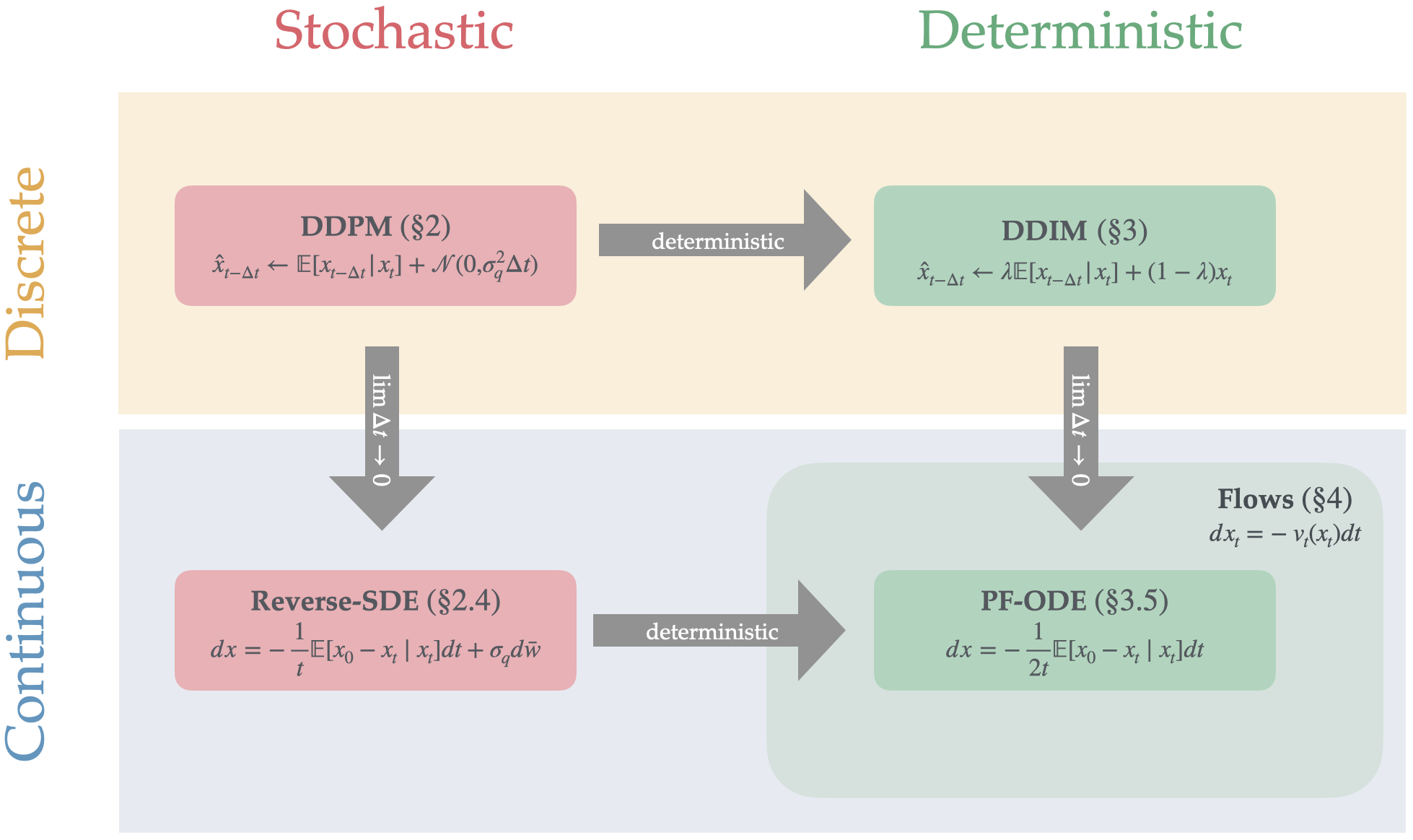}
    \caption{Commutative diagram of the different reverse samplers described in this tutorial, and their relations.
    Each deterministic sampler produces identical marginal distributions as its stochastic counterpart.
    There are also various ways to construct stochastic versions of flows,
    which are not pictured here (e.g. \citet{albergo2023stochastic}).
    } 
    \label{fig:concept-map}
\end{figure*}

\begin{itemize}
\item See Figure~\ref{fig:concept-map} for a diagram of the 
different methods described in this tutorial, and their relations.
    \item We highly recommend the flow-matching tutorial of \citet{mathieu2024flow},
    which includes helpful visualizations of flows, and uses notation
    more consistent with the current literature.
    \item As a curiosity, note that we never had to define an explicit ``forward process'' for flow-matching,
    as we did for Gaussian diffusion. Rather, it was enough to
    define the appropriate ``reverse processes'' (via flows).
    \item What we called \emph{pointwise flows} are also called
    \emph{two-sided conditional flows} in the literature, and was developed in
    \citet{albergo2022building,pooladian2023multisample,liu2022flow,tong2023simulation}.
    \item \citet{albergo2023stochastic} define the framework of \emph{stochastic interpolants},
    which can be thought of as considering stochastic pointwise flows, instead of only deterministic ones.
    Their framework strictly generalizes both DDPM and DDIM.
    \item See \citet{stark2024dirichlet} 
    for an interesting example of non-standard flows.
    They derive a generative model for discrete spaces by embedding
    into a continuous space (the probability simplex), then 
    constructing a special flow on these simplices.
\end{itemize}

\clearpage
\newpage
\section{Diffusion in Practice}
\label{sec:practical}

To conclude, we mention some aspects of diffusion which are important in practice,
but were not covered in this tutorial.

\paragraph{Samplers in Practice.}
Our DDPM and DDIM samplers (algorithms \ref{alg:ddpm_gen} and \ref{alg:ddim_inf}) correspond to the samplers presented in \citet{ho2020denoising} and \citet{song2021denoising}, respectively, but with different choice of schedule and parametrization (see footnote \ref{foot:ddpm_diff}). DDPM and DDIM were some of the earliest samplers to be used in practice, but since then there has been significant progress in samplers for fewer-step generation (which is crucial since each step requires a typically-expensive model forward-pass).\footnote{Even the best samplers still require around $10$ sampling steps, which may be impractical. A variety of \emph{time distillation} methods seek to train one-step-generator student models to match the output of diffusion teacher models, with the goal of high-quality sampling in one (or few) steps. Some examples include consistency models \citep{song2023consistency} and adversarial distillation methods \citep{lin2024sdxl, xu2023ufogen, sauer2024fast}. Note, however, that the distilled models are no longer diffusion models, nor are their samplers (even if multi-step) diffusion samplers.}
In sections \ref{sec:sdes} and \ref{sec:pf_ode}, we showed that DDPM and DDIM can be seen as discretizations of the reverse SDE and Probability Flow ODE, respectively. The SDE and ODE perspectives automatically lead to many samplers corresponding to different black-box SDE and ODE numerical solvers (such as Euler, Heun, and Runge-Kutta).
It is also possible to take advantage of the specific
structure of the diffusion ODE, to improve upon black-box solvers
\citep{lu2022dpm,lu2022dpmplusplus,zhang2023fast}.

\paragraph{Noise Schedules.}
The \emph{noise schedule} typically refers to $\sigma_t$, which determines the amount of noise added at time $t$ of the diffusion process. 
The simple diffusion (\ref{eqn:intro-fwd}) has $p(x_t) \sim \cN(x_0, \sigma_t^2)$ with $\sigma_t \propto \sqrt{t}$. Notice that the variance of $x_t$ increases at every timestep.\footnote{\citet{song2020score} made the distinction between ``variance-exploding'' (VE) and ``variance-preserving'' (VP) schedules while comparing SMLD \citep{song2019generative} and DDPM \citep{ho2020denoising}. The terms VE and VP often refer specifically to SMLD and DDPM, respectively. Our diffusion \eqref{eqn:intro-fwd} could also be called a variance-exploding schedule, though our noise schedule differs from the one originally proposed in \citet{song2019generative}.}

In practice, schedules with controlled variance are often preferred. One of the most popular schedules, introduced in \citet{ho2020denoising}, uses a time-dependent variance and scaling such that the variance of $x_t$ remains bounded. 
Their discrete update is
\begin{align}
x_t &= \sqrt{1-\beta(t)} x_{t-1} + \sqrt{\beta(t)} \epsilon_t; \quad \epsilon_t \sim \cN(0, 1),
\label{eq:ho_update}
\end{align}
where $0 < \beta(t)< 1$ is chosen so that $x_t$ is (very close to) clean data at $t=1$ and pure noise at $t=T$.

The general SDE (\ref{eq:general_sde}) introduced in \ref{sec:sdes} offers additional flexibility. Our simple diffusion (\ref{eqn:intro-fwd}) has $f = 0$, $g = \sigma_q$, while the diffusion \eqref{eq:ho_update} of \citet{ho2020denoising} has $f = -\frac{1}{2} \beta(t)$,  $g = \sqrt{\beta(t)}$. \citet{karras2022elucidating} reparametrize the SDE in terms of an overall scaling $s(t)$ and variance $\sigma(t)$ of $x_t$, as a more interpretable way to think about diffusion designs, and suggest a schedule with $s(t) = 1$, $\sigma(t) = t$ (which corresponds to $f=0$, $g = \sqrt{2t}$). Generally, the choice of $f, g$, or equivalently $s, \sigma$, offers a convenient way to explore the design-space of possible schedules.

\paragraph{Likelihood Interpretations and VAEs.}
One popular and useful interpretation of diffusion models
is the Variational Auto Encoder (VAE) perspective\footnote{This was actually the original approach to derive the diffusion objective function, in \citet{OGdiffusion} and also \citet{ho2020denoising}.}.
Briefly, diffusion models can be viewed as a special
case of a deep hierarchical VAE, where each diffusion
timestep corresponds to one ``layer'' of the VAE decoder.
The corresponding VAE encoder is given by the
forward diffusion process, which produces the sequence
of noisy $\{x_t\}$ as the ``latents'' for input $x$.
Notably, the VAE encoder here is not learnt, unlike
usual VAEs.
Because of the Markovian structure of the latents,
each layer of the VAE decoder can be trained in isolation,
without forward/backward passing through all previous layers;
this helps with the notorious training
instability of deep VAEs.
We recommend the tutorials of
\citet{turner_blog} and
\citet{luo2022understanding}
for more details on the VAE perspective.

One advantage of the VAE interpretation is, it gives
us an estimate of the \emph{data likelihood} under our generative model,
by using the standard Evidence-Based-Lower-Bound (ELBO) for VAEs.
This allows us to train diffusion models directly using a maximum-likelihood objective.
It turns out that the ELBO for the diffusion VAE
reduces to exactly the L2 regression loss 
that we presented, but with a particular
\emph{time-weighting} that weights 
the regression loss differently at different time-steps $t$.
For example, regression errors at large times $t$ (i.e. at high noise levels)
may need to be weighted differently from errors at small times,
in order for the overall loss to properly reflect a likelihood.\footnote{
See also Equation (5) in \citet{kadkhodaie2024generalization}
for a simple bound on KL divergence between
the true distribution and generated distribution,
in terms of regression excess risks.}
The best choice of time-weighting in practice, however,
is still up for debate: the ``principled'' choice informed by the 
VAE interpretation does not always produce the best generated samples\footnote{
For example, \citet{ho2020denoising} drops the time-weighting terms, and just uniformly
weights all timesteps.}.
See \citet{kingma2023understanding} for a good discussion of different weightings and their effect.

\paragraph{Parametrization: $x_0$ / $\epsilon$ / v -prediction.}
Another important practical choice is which of several closely-related quantities -- partially-denoised data, fully-denoised data, or the noise itself -- we ask the network to predict.\footnote{More accurately, the network always predicts conditional expectations of these quantities.} Recall that in DDPM Training (Algorithm \ref{alg:ddpm_train}), we asked the network $f_\theta$ to learn to predict $\E[x_{t-\dt} | x_t]$ by minimizing $\|f_\theta(x_t, t) - x_{t-\dt}\|_2^2$. However, other parametrizations are possible. For example, recalling that $\E[x_{t-\dt} - x_t | x_t] \overset{\text{eq. } \ref{eq:var_reduc_dt}}{=} \frac{\dt}{t} \E[x_0 - x_t | x_t]$, we see that that
\begin{align*}
    \underset{\theta}{\min} &\quad \|f_\theta(x_t, t) - x_0\|_2^2 \implies f^\star_\theta(x_t, t) = \E[x_0 | x_t]
\end{align*}
is a (nearly) equivalent problem, which is often called \emph{$x_0$-prediction}.\footnote{This corresponds to the variance-reduced algorithm (\ref{alg:ddpm_train_varred}).} The objectives differ only by a time-weighting factor of $\frac{1}{t}$. Similarly, defining the \emph{noise} $\epsilon_t = \frac{1}{\sigma_t} \E[x_0 - x_t | x_t]$, we see that we could alternatively ask the the network to predict $\E[\epsilon_t | x_t]$: this is usually called \emph{$\epsilon$-prediction}. Another parametrization, \emph{v-prediction}, asks the model to predict $v = \alpha_t \epsilon - \sigma_t x_0$ \citep{salimans2022progressive} -- mostly predicting data for high noise-levels and mostly noise for low noise-levels. All the parametrizations differ only by time-weightings (see Appendix \ref{append:x0_vs_eps_param} for more details).

Although the different time-weightings do not affect the optimal solution, they do impact training as discussed above. Furthermore, even if the time-weightings are adjusted to yield equivalent problems in principle, the different parametrizations may behave differently in practice, since learning is not perfect and certain objectives may be more robust to error. For example, $x_0$-prediction combined with a schedule that places a lot of weight on low noise levels may not work well in practice, since for low noise the identity function can achieve a relatively low objective value, but clearly is not what we want.

\paragraph{Sources of Error.}
Finally, when using diffusion and flow models in practice, there are a number of
sources of error which prevent the learnt generative model from exactly producing the target distribution.
These can be roughly segregated into training-time and sampling-time errors.
\begin{enumerate}
    \item Train-time error: Regression errors in learning the population-optimal
    regression function.
    The regression objective is the marginal flow $v^*_t$ in flow-matching,
    or the scores $\E[x_0 \ \mid x_t]$ in diffusion models.
    For each fixed time $t$, this a standard kind of statistical error. It depends on the neural network architecture and size as well as the number of samples,
    and can be decomposed further into approximation and estimation errors in the usual way (e.g. see \citet[Sec. 4]{advani2020high} decomposing a 2-layer network into approximation error and over-fitting error).

    \item Sampling-time error:
    Discretization errors from using finite step-sizes $\dt$.
    This error is exactly the discretization error of the ODE or SDE solver
    used in sampling. These errors manifest in different ways: for DDPM, this reflects
    the error in using a Gaussian approximation of the reverse process (i.e. Fact 1 breaks for large $\sigma$).
    For DDIM and flow matching, it reflects the error in simulating continuous-time flows in discrete time.
\end{enumerate}
These errors interact and compound in nontrivial ways, which are not yet fully understood.
For example, it is not clear exactly how train-time error in the regression estimates
translates into distributional error of the entire generative model.
(And this question itself is complicated, since it is not always clear what type of distributional divergence
we care about in practice). 
Interestingly, these ``errors'' can also have a beneficial effect 
on small train sets,
because they act as a kind of \emph{regularization} which prevents the diffusion model
from just memorizing the train samples (as discussed in Section~\ref{sec:generalization}).

\subsection*{Conclusion}
We have now covered the basics of diffusion models and flow matching.
This is an active area of research, and
there are many interesting aspects and open questions which 
we did not cover (see Page~\pageref{sec:rec_reading} for recommended reading).
We hope the foundations here equip the reader to
understand more advanced topics in diffusion modeling,
and perhaps contribute to the research themselves.

\vfill
\vfill

\appendix
\newpage
\begin{fullwidth}
\section{Additional Resources}
\label{sec:rec_reading}

Several other helpful resources
for learning diffusion (tutorials, blogs, papers), roughly in order of mathematical background required.

\newcommand{\Tutorial}[4]{
  \item \textbf{#1} \\
  #2. (#4) \\
  \begin{quote}
   #3 
  \end{quote}
}

\begin{enumerate}[leftmargin=*]
\Tutorial{Perspectives on diffusion.}{\citet*{dieleman2023perspectives}}{Overview of many interpretations of diffusion, and techniques.}{Webpage.}

\Tutorial{Tutorial on Diffusion Models for Imaging and Vision.}{\citet*{chan2024tutorial}}{More focus on intuitions and applications.}{49 pgs.}

\Tutorial{Interpreting and improving diffusion models using the euclidean distance function.}{\citet*{permenter2023interpreting}}{Distance-field interpretation.
See accompanying blog with simple code \citep*{CYblog}.}{Webpage.}

\Tutorial{On the Mathematics of Diffusion Models.}{\citet*{mcallester2023mathematics}}{Short and accessible.}{4 pgs.}

\Tutorial{Building Diffusion Model's theory from ground up}{\citet*{das2024buildingdiffusionmodels}}{ICLR 2024 Blogposts Track.
Focus on SDE and score-matching perspective.}{Webpage.}

\Tutorial{Denoising Diffusion Models: A Generative Learning Big Bang.}{\citet*{cvpr2023}}{CVPR 2023 tutorial, with recording.}{Video, 3 hrs.}

\Tutorial{Diffusion Models From Scratch.}{\citet*{tonydiff}}{Fairly complete on topics, includes: DDPM, DDIM, \citet{karras2022elucidating}, SDE/ODE solvers. Includes practical remarks and code.}{Webpage, 10 parts.}

\Tutorial{Understanding Diffusion Models: A Unified Perspective.}{\citet*{luo2022understanding}}{Focus on VAE interpretation, with explicit math details.}{22 pgs.}

\Tutorial{Demystifying Variational Diffusion Models.}{\citet*{ribeiro2024demystifying}}{Focus on VAE interpretation, with explicit math details.}{44 pgs.}

\Tutorial{Diffusion and Score-Based Generative Models.}{\citet*{songVideo}}{Discusses several interpretations, applications, and comparisons to other generative modeling methods.}{Video, 1.5 hrs.}

\Tutorial{Deep Unsupervised Learning using Nonequilibrium Thermodynamics}{\citet*{OGdiffusion}}{Original paper introducing diffusion models for ML. Includes unified description of discrete diffusion (i.e. diffusion on discrete state spaces).}{9 pgs + Appendix}

\Tutorial{An Introduction to Flow Matching.}{\citet*{mathieu2024flow}}{Insightful figures and animations, with rigorous mathematical exposition.}{Webpage.}

\Tutorial{Elucidating the Design Space of Diffusion-Based Generative Models.}{\citet*{karras2022elucidating}}{Discusses the effect of various design choices such as noise schhedule, parameterization, ODE solver, etc. Presents a generalized framework that captures many choices.}{10 pgs + Appendix.}

\Tutorial{Denoising Diffusion Models}{\citet*{peyre2023denoising}}{Fast-track through the mathematics, for readers already comfortable with Langevin dynamics and SDEs.}{4 pgs.}

\Tutorial{Generative Modeling by Estimating Gradients of the Data Distribution.}{\citet*{song2020score}}{Presents the connections between SDEs, ODEs, DDIM, and DDPM.}{9 pgs + Appendix.}

\Tutorial{Stochastic Interpolants: A Unifying Framework for Flows and Diffusions.}{\citet*{albergo2023stochastic}}{Presents a general framework that captures many diffusion variants, and learning objectives. For readers comfortable with SDEs }{46 pgs + Appendix.}

\Tutorial{Sampling, Diffusions, and Stochastic Localization.}{\citet*{montanari2023sampling}}{Presents diffusion as a special case of ``stochastic localization,'' a technique used in high-dimensional statistics to establish mixing of Markov chains.}{22 pgs + Appendix.}
 
\end{enumerate}

\end{fullwidth}

\newpage
\section{Omitted Derivations}
\subsection{KL Error in Gaussian Approximation of Reverse Process}
\label{app:kl}

Here we prove Lemma~\ref{lem:KL}, restated below.

\begin{lemma}
\label{lem:KL-restated}
Let $p(x)$ be an arbitrary density over $\R$,
with bounded 1st to 4th order derivatives.
Consider the joint distribution $(x_0, x_1)$,
where $x_0 \sim p$ and $x_1 \sim x_0 + \cN(0, \sigma^2)$.
Then, for any conditioning $z \in \R$
we have
\[
\mathrm{KL}\left(
\cN(\mu_z, \sigma^2 ) 
~~||~~
p_{x_0 \mid x_1}(\cdot \mid x_1 = z)
\right)
\leq O(\sigma^4)
\]
where
\[
\mu_z := z + \sigma^2 \nabla \log p(z).
\]
\end{lemma}

\begin{fullwidth}
\begin{proof}
WLOG, we can take $z=0$. 
We want to estimate the KL:
\begin{align}
KL(  \cN(\mu, \sigma^2) || p(x_0 = \cdot \mid x_1 = 0) ) 
\end{align}
where we will let $\mu$ be arbitrary for now.

Let $q := \cN(\mu, \sigma^2)$,
and $p(x) =: \exp(F(x))$.
We have $x_1 \sim p \star \cN(0, \sigma^2)$.
This implies:
\[
p(x_1 = x) = \E_{\eta \sim \cN(0, \sigma^2)}[ p(x + \eta) ].
\]

Let us first expand the logs of the two distributions we are comparing:
\begin{align}
&\log p(x_0 = x \mid x_1 = 0)   \\
&= 
\log p(x_1 = 0 | x_0 = x) + \log p(x_0 = x) - \log p(x_1 = 0) \\
&=
-\log(\sigma \sqrt{2\pi})
-0.5 x^2 \sigma^{-2}
+ \log p(x_0 = x) - \log p(x_1 = 0) \\
&=
-\log(\sigma \sqrt{2\pi})
-0.5 x^2 \sigma^{-2}
+ F(x) - \log p(x_1 = 0) \\
\end{align}

And also:
\begin{align}
\log q(x) &=
-\log(\sigma \sqrt{2\pi})
-0.5 (x-\mu)^2 \sigma^{-2}
\end{align}

Now we can expand the KL:
\begin{align}
&KL( q || p(x_0 = \cdot \mid x_1 = 0))\\
&= 
\E_{x \sim q}[
\log q(x) - \log p(x_0 = x \mid x_1 = 0)
]\\
&= 
\E_{x \sim q}[
-\log(\sigma \sqrt{2\pi})
-0.5 (x-\mu)^2 \sigma^{-2}
-(-\log(\sigma \sqrt{2\pi})
-0.5 x^2 \sigma^{-2}
+ F(x) - \log p(x_1 = 0))
]\\
&= 
\E_{x \sim q}[
-0.5 (x-\mu)^2 \sigma^{-2}
+0.5 x^2 \sigma^{-2}
- F(x) + \log p(x_1 = 0))
]\\
&= 
\E_{\eta \sim \cN(0, \sigma^2); x = \mu+\eta}[
-0.5 \eta^2 \sigma^{-2}
+0.5 x^2 \sigma^{-2}
- F(x) + \log p(x_1 = 0))
]
\tag{work}
\\
&= 
-0.5 \E[\eta^2] \sigma^{-2}
+0.5 \E[x^2] \sigma^{-2}
- \E[F(x)] + \log p(x_1 = 0))
]\\
&= 
-0.5 \sigma^2 \sigma^{-2}
+0.5 (\sigma^2 + \mu^2) \sigma^{-2}
- \E[F(x)]
+ \log p(x_1 = 0))
]\\
&=
0.5 \mu^2 \sigma^{-2}
+ \log p(x_1 = 0) 
- \E_{x\sim q}[F(x)] \\
&\approx
0.5 \mu^2 \sigma^{-2}
+ \log p(x_1 = 0)
- \E_{x \sim q}[
F(0) + F'(0)x + 0.5F''(0)x^2 + O(x^3) + O(x^4)] \\
&= \label{ln:pp}
\log p(x_1 = 0)
+ 0.5 \mu^2 \sigma^{-2}
- F(0)
- F'(0)\mu - 0.5F''(0)(\mu^2 + \sigma^2) + O(\sigma^2\mu + \mu^2 + \sigma^4) 
\end{align}

We will now estimate the first term, $\log p(x_1=0)$:
\begin{align}
&\log p(x_1 = 0) \\
&= \log \E_{\eta \sim \cN(0, \sigma^2)}[p(\eta)] \\
&= \log \E_{\eta \sim \cN(0, \sigma^2)}[p(0) + p'(0)\eta + 0.5p''(0)\eta^2 + O(\eta^3) + O(\eta^4)] \\
&= \log \left( p(0) + 0.5p''(0)\sigma^2 + O(\sigma^4) \right)\\
&= \log p(0)  + \frac{0.5p''(0)\sigma^2 + O(\sigma^4)}{p(0)} + O( \sigma^4 )
\tag{Taylor expand $\log(p(0) + \eps)$ around $p(0)$} \\
\label{ln:logp1}
&= \log p(0)  + 0.5\sigma^2\frac{p''(0)}{p(0)} + O( \sigma^4 )
\end{align}
To compute the derivatives of $p$, observe that:
\begin{align}
F(x) &= \log p(x) \\
\implies 
F'(x) &= p'(x)/p(x) \\
\implies
F''(x) &= p''(x)/p(x) - (p'(x)/p(x))^2 \\
&= p''(x)/p(x) - (F'(x))^2 \\
\implies 
p''(x)/p(x)
&= F''(x) + (F'(x))^2
\label{ln:ppp}
\end{align}

Thus, continuing from line~\eqref{ln:logp1}:
\begin{align}    
\log p(x_1 = 0) &= \log p(0)  + 0.5\sigma^2\frac{p''(0)}{p(0)} + O( \sigma^4 )\\
&= F(0)  + 0.5\sigma^2(F''(0) - F'(0)^2)  + O( \sigma^4 ) \tag{by Line~\ref{ln:ppp}}
\end{align}

We can now plug this estimate of $\log p(x_1=0)$ into Line~\eqref{ln:pp}.
We omit the argument $(0)$ from $F$ for simplicity:
\begin{align}
&KL( q || p(x_0 = \cdot \mid x_1 = 0))\\
&=\boxed{\log p(x_1 = 0)}
+ 0.5 \mu^2 \sigma^{-2}
- F
- F'\mu - 0.5F''(\mu^2 + \sigma^2) + O(\mu^4 + \sigma^4)  \\
&=
F + 0.5\sigma^2(F'' + F'^2)
+ 0.5 \mu^2 \sigma^{-2}
- F
- F'\mu - 0.5F''(\mu^2 + \sigma^2) + O(\mu^4 + \sigma^4)  \\
&=
+ 0.5\sigma^2F''
+ 0.5\sigma^2F'^2
+ 0.5 \mu^2 \sigma^{-2}
- F'\mu - 0.5F''\mu^2 
-0.5F''\sigma^2
+ O(\mu^4 + \sigma^4)   \\
&=
- F'\mu 
+ 0.5 \mu^2 \sigma^{-2}
+ 0.5F'^2\sigma^2
- 0.5F''\mu^2 
+ O(\mu^4 + \sigma^4) 
\end{align}

Up to this point, $\mu$ was arbitrary. We now set
\[
\mu_* := F'(0)\sigma^2.
\]

And continue:
\begin{align}
&KL( q || p(x_0 = \cdot \mid x_1 = 0))\\
&=
- F'\mu_*
+ 0.5 \mu_*^2 \sigma^{-2}
+ 0.5F'^2\sigma^2
- 0.5F''\mu_*^2 
+ O(\mu_*^4 + \sigma^4)\\
&=
- F'^2\sigma^2
+ 0.5 F'^2 \sigma^2
+ 0.5F'^2\sigma^2
+ O(\sigma^4) \label{line:cancel} \\
&= O(\sigma^4)
\end{align}
as desired.

\end{proof}
Notice that our choice of $\mu_*$ in the above proof
was crucial; for example if we had set $\mu_*=0$, 
the $\Omega(\sigma^2)$ terms in Line~\eqref{line:cancel} would not have cancelled out.

\end{fullwidth}
\subsection{SDE proof sketches}
\label{append:sde}

Here is sketch of the proof of the equivalence of the SDE and Probability Flow ODE, which relies on the equivalence of the SDE to a Fokker-Planck equation. (See \citet{song2020score} for full proof.)
\begin{proof}
    \begin{align*}
    dx &= f(x,t)dt + g(t)dw \\
    \iff \frac{\partial p_t(x)}{\partial t} &= -\nabla_x (f p_t) + \frac{1}{2} g^2 \nabla_x^2 p_t \quad \text{(FP)} \\
    &= -\nabla_x (f p_t) + \frac{1}{2} g^2 \nabla_x (p_t \nabla_x \log p_t) \\
    &= -\nabla_x \{ (f - \frac{1}{2} g^2 \nabla_x \log p_t) p_t \} \\
    &= -\nabla_x \{ \tilde f(x, t) p_t(x) \}, \quad \tilde f(x, t) = f(x,t) - \frac{1}{2} g(t)^2 \nabla_x \log p_t(x) \\
    \implies dx &= \tilde f(x, t) dt
\end{align*}
\end{proof}

The equivalence of the SDE and Fokker-Planck equations follows from It\^{o}'s formula and integration-by-parts. Here is an outline for a simplified case in 1d, where $g$ is constant (see \citet{winkler_blog_fp} for full proof):
\begin{proof}
    \begin{align*}
    dx &= f(x) dt + g dw, \quad dw \sim \sqrt{dt} \mathcal{N}(0,1) \\
    \text{For any $\phi$:} \quad d\phi(x) 
    &= \biggr( f(x) \partial_x \phi(x) + \frac{1}{2} g^2 \partial_x^2 \phi(x) \biggr) dt + g \partial_x \phi(x) dw \mathnote{It\^{o}'s formula} \\
    \implies \frac{d}{dt} \E[\phi] &= \E[f \partial_x \phi + \frac{1}{2} g^2 \partial_x^2 \phi], \quad (\E[dw] = 0) \\ 
    \int \phi(x) \partial_t p(x,t) dx &= \int f(x) \partial_x \phi(x) p(x,t) dx + \frac{1}{2} g^2 \int \partial_x^2 \phi(x) p(x,t) dx \\
    &= - \int \phi(x) \partial_x (f(x) p(x,t)) dx + \frac{1}{2} g^2 \int \phi(x) \partial_x^2 p(x,t) dx, \mathnote{integration-by-parts} \\
    \partial_t p(x) &= - \partial_x (f(x) p(x,t)) + \frac{1}{2} g^2 \partial_{x}^2 p(x), \mathnote{Fokker-Planck} \\
    \end{align*}
\end{proof}

\subsection{DDIM Point-mass Claim}
\label{app:ddim}

Here is a version of Claim~\ref{claim:ddim_one_pt} 
where $p_0$ is a delta at an arbitrary point $x_0$.
\begin{claim}
\label{claim:ddim_x0}
    Suppose the target distribution is a point mass
    at $x_0 \in \R^d$, i.e. $p_0 = \delta_{x_0}$.
    Define the function
\begin{align}
   \rG(x_t) = \left(\frac{\sigma_{t-\dt}}{\sigma_t}\right)(x_t - x_0) + x_0.
\end{align}
Then we clearly have $\rG \sharp p_t = p_{t-\dt}$, and moreover
    \[
    \label{eqn:FtExp}
    \rG(x_t)
    = 
    x_t + \lambda(\E[x_{t - \dt} \mid x_t] - x_t)
    =: F_t(x_t).
    \]
Thus Algorithm 2 defines a valid reverse sampler for target distribution
$p_0 = \delta_{x_0}.$
\end{claim}

\subsection{Flow Combining Lemma}
\label{sec:flow_formal}
Here we provide a more formal
statement of the marginal flow result stated in Equation (\ref{eqn:vexp}).

Equation~\eqref{eqn:vexp} follows from a more general lemma (Lemma~\ref{lem:master-flow})
which formalizes the ``gas combination'' analogy of Section~\ref{sec:ddim}.
The motivation for this lemma is, we need a way of combining flows:
of taking several different flows and producing a single ``effective flow.''
As a warm-up for the lemma, suppose we have $n$ different flows, each with their own 
initial and final distributions $q_i, p_i$:
\begin{align*}
q_1 \flowto{v^{(1)}} p_1, \quad
q_2 \flowto{v^{(2)}} p_2, \quad
\ldots, \quad
q_n \flowto{v^{(n)}} p_n
\end{align*}
We can imagine these as the flow of $n$ different gases,
where gas $i$ has initial density $q_i$ and final density $p_i$.
Now we want to construct an overall flow $v^*$
which takes the average initial-density to the average final-density:
\[
\E_{i \in [n]}[q_i]
\flowto{v^*}
\E_{i \in [n]}[p_i].
\]
To construct $v^*_t(x_t)$, we must take an average of the individual
vector fields $v^{(i)}$, weighted by the 
probability mass the $i$-th flow places on $x_t$, at time $t$.
(This is exactly analogous to Figure~\ref{fig:gas}).

This construction is formalized in Lemma~\ref{lem:master-flow}.
There, instead of averaging over just a finite set of flows, we are allowed
to average over any \emph{distribution} over flows.
To recover Equation~\eqref{eqn:vexp}, we can apply Lemma~\ref{lem:master-flow}
to a distribution $\Gamma$ over
$(v, q_v) = (\Tab{x_1}{x_0}, \delta_{x_1})$,
that is, pointwise flows and their associated initial delta distributions.

\begin{mdframed}[nobreak]
\begin{lemma}[Flow Combining Lemma]
\label{lem:master-flow}
Let $\Gamma$ be an arbitrary joint distribution over 
pairs $(v, q_v)$ of flows $v$ and their associated initial
distributions $q_v$.
Let $v(q_v)$ denote the final distribution when initial distribution $q_v$
is transported by flow $v$, so $q_v \flowto{v} v(q_v)$

For fixed $t \in [0, 1]$, consider the joint distribution over
$(x_1, x_t, w_t) \in (\R^d)^3$ generated by:
\begin{align*}
(v, q_v) &\sim \Gamma \\
x_1 &\sim q_v \\
x_t &:= \mathrm{RunFlow}(v, x_1, t) \\
w_t &:= v_t(x_t).
\end{align*}
Then, taking all expectations w.r.t. this joint distribution, the flow $v^*$ defined as
\begin{align}
v^*_t(x_t) &:= \E[w_t \mid x_t] \\
&= \E[v_t(x_t) \mid x_t] 
\end{align}
is known as the \emph{marginal flow for $\Gamma$}, and transports:
\begin{align}
\E[q_v]
\flowto{v^*}
\E[ v(q_v) ].
\end{align}

\end{lemma}
\end{mdframed}

\subsection{Derivation of DDIM using Flows}
\label{app:ddim_pf_by_flow}
Now that we have the machinery of flows in hand, it is fairly easy to 
derive the DDIM algorithm ``from scratch'', by extending our simple scaling algorithm
from the single point-mass case.

First, we need to find the pointwise flow. Recall from Claim \ref{claim:ddim_x0} that for the simple case where the target distribution $p_0$ is a Dirac-delta at $x_0$, the following scaling maps $p_t$ to $p_{t-\dt}$:
\begin{align*}
    \rG(x_t) = \left(\frac{\sigma_{t-\dt}}{\sigma_t}\right)(x_t - x_0) + x_0 
    \implies 
    G_t \sharp p_t = p_{t-\dt}.
\end{align*}
$G_t$ implies the pointwise flow:
\begin{align*}
    \lim_{t \to 0} \quad \left(\frac{\sigma_{t-\dt}}{\sigma_t}\right)  
    &= \sqrt{1 - \frac{\dt}{t}} = (1 - \frac{\dt}{2t}) \\
    \implies \Tab{x_1}{x_0}_t(x_t) &= -\underset{\dt \to 0}{\lim} \frac{G_t(x_t) - x_t}{\dt} 
    = \frac{1}{2t} (x_t - x_0),
\end{align*}
which agrees with~\eqref{eqn:ptwise_ddim}.

Now let us compute the marginal flow $v^*$ generated by the pointwise flow of Equation~\eqref{eqn:ptwise_ddim}
and the coupling implied by the diffusion forward process.
By Equation \eqref{eqn:vstar}, the marginal flow is:
\begin{align*}
v^*_t(x_t)
&= \E_{x_1, x_0 \mid x_t}[ \Tab{x_1}{x_0}_t(x_t) \mid x_t ] \mathnote{By gas-lemma.}\\
&= \frac{1}{2t} \E_{
    \substack{x_0 \sim p;~ x_1 \gets x_0 + \cN(0, \sigma_q^2) \\
    x_t \gets \mathrm{RunFlow}(\Tab{x_1}{x_0}_t, x_1, t)}}
[ x_0 - x_t \mid x_t ] \mathnote{For our choices of coupling and flow.}\\
&= \frac{1}{2t} \E_{
    \substack{x_0 \sim p;~ x_1 \gets x_0 + \cN(0, \sigma_q^2) \\
    x_t \gets x_1 \sqrt{t} + (1-\sqrt{t})x_0
    }}[ x_0 - x_t \mid x_t ] \mathnote{Expanding the flow trajectory.} \\
&= \frac{1}{2t} \E_{
    \substack{x_0 \sim p\\
    x_t \gets \sqrt{t} \cN(0, \sigma_q^2)
    }}[ x_0 - x_t \mid x_t ] \mathnote{Plugging in $x_1 = x_0 + \cN(0, \sigma_q^2).$}
\end{align*}
This is exactly the differential equation describing the
trajectory of DDIM
(see Equation~\ref{eq:simple_pf_ode}, which is the continuous-time
limit of Equation~\ref{eqn:ddim}).

\subsection{Two Pointwise Flows for DDIM give the same Trajectory}
\label{app:ddim_two_ptwise_flows_equiv}

We want to show that pointwise flow~\ref{eqn:ptwise_ddim2}:
\[
\Tab{x_1}{x_0}_t(x_t) = \frac{1}{2\sqrt{t}}(x_0 - x_1)
\]
is equivalent to the DDIM pointwise flow~\eqref{eqn:ptwise_ddim}:
\[
\Tab{x_1}{x_0}_t(x_t)= \frac{1}{2t} (x_t - x_0)
\]
because both these pointwise flows generate the same trajectory of $x_t$:
\[
x_t = x_0 + (x_1 - x_0)\sqrt{t}.
\]
To see this, we can solve the ODE determined by~\eqref{eqn:ptwise_ddim} via the Separable Equations method:
\begin{align*}
    \frac{dx_t}{dt} &= -\frac{1}{2t} (x_0 - x_t) \\
    \implies \frac{\frac{dx_t}{dt}}{x_t - x_0} &= \frac{1}{2t} \\
    \implies \int \frac{1}{x_t - x_0} dx &= \int \frac{1}{2t} dt, 
    \text{ since } \frac{dx_t}{dt} dt = dx \\
    \implies \log(x_t - x_0) &= \log \sqrt{t} + c \\
    c &= \log(x_1 - x_0) \text{ (boundary cond.)}\\
    \implies \log(x_t - x_0) &= \log \sqrt{t} (x_1 - x_0)  \\
    \implies x_t - x_0 &= \sqrt{t} (x_1 - x_0).
\end{align*}

\subsection{DDIM vs Time-reparameterized linear flows}
\label{app:ddim_sqrt_time}

\begin{lemma}[DDIM vs Linear Flows]
\newcommand{\pddim}{p^{\textrm{ddim}}}

Let $p_0$ be an arbitrary target distribution.
Let $\{x_t\}_t$ be the joint distribution defined by the
DDPM forward process applied to $p_0$, so the marginal distribution of $x_t$
is $p_t = p \star \cN(0, t\sigma_q^2)$.

Let $x^* \in \R^d$ be an arbitrary initial point.
Consider the following two deterministic trajectories:
\begin{enumerate}
    \item The trajectory $\{y_t\}_t$ of the continuous-time DDIM flow,
    with respect to target distribution $p_0$,
    when started at initial point $y_1 = x^*$.
    
    That is, $y_t$ is the solution to the following ODE (Equation~\ref{eq:simple_pf_ode}):
    \begin{align}
    \frac{dy_t}{dt} &= -v^{\mathrm{ddim}}(y_t) \\
    &= -\frac{1}{2t} \E_{x_0 \mid x_t}[x_0 - x_t \mid x_t=y_t] 
    \end{align}
    with boundary condition $y_1$ at $t=1$.

    \item The trajectory $\{z_t\}_t$ 
    produced when initial point $z_1 = x^*$ is transported
    by the marginal flow constructed from:
    \begin{itemize}
        \item Linear pointwise flows
        \item The DDPM-coupling of Line~\eqref{eqn:ddpm_coupling}.
    \end{itemize}
    That is, the marginal flow
    \begin{align*}
    v^\star_t(x_t) &= \E_{x_0, x_1|x_t} [\Tab{x_1}{x_0}(x_t) | x_t] \\
    &:= \E_{x_0, x_1|x_t} [x_0 - x_1 | x_t] \\
    &= \E_{x_0|x_t} [x_0 - x_t | x_t] \mathnote{since \mbox{$\E[x_1 | x_t] = x_t$} under the DDPM coupling.} 
\end{align*}

\end{enumerate}
Then, we claim these two trajectories are identical with the following time-reparameterization:
\begin{align}
\forall t \in [0,1]: \quad
y_t = z_{\sqrt{t}}
\end{align}
\end{lemma}

\subsection{Proof Sketch of Claim~\ref{claim:var_red}}
\label{app:var_proof}
We will show that, in the forward diffusion setup of Section~\ref{sec:fundamentals}:
\begin{align}
\E[(x_{t} - x_{t-\dt}) \mid x_t]  
= 
\frac{\dt}{t}
\E[(x_t - x_0) \mid x_t].
\end{align}

\begin{proof}[Proof sketch]

Recall $\eta_t = x_{t+\dt} - x_t$. So by linearity of expectation:
\begin{align}
\E[(x_t - x_0) \mid x_t]
&=
\E[\sum_{i<t} \eta_i \mid x_t] \\
&=
\sum_{i < t} \E[\eta_i \mid x_t].
\label{ln:app_v}
\end{align}
Now, we claim that for given $x_t$,
the conditional distributions
$p(\eta_i \mid x_t)$ are identical for all $i < t$.
To see this, notice that the joint distribution function
$p(x_0, x_t, \eta_0, \eta_\dt, \dots, \eta_{t-\dt})$ is symmetric in the $\{\eta_i\}$s,
by definition of the forward process,
and therefore
the conditional distribution function
$p(\eta_0, \eta_\dt, \dots, \eta_{t-\dt} \mid x_t)$ is also symmetric in the $\{\eta_i\}$s.
Therefore, all $\eta_i$ have identical conditional expectations:
\[
\E[\eta_0 \mid x_t] 
= \E[\eta_\dt \mid x_t] 
= \dots
= \E[\eta_{t-\dt} \mid x_t]
\]
And since there are $(t/\dt)$ of them,
\[
\sum_{i < t} \E[\eta_i \mid x_t]
= 
\frac{t}{\dt} \E[\eta_{t-\dt} \mid x_t].
\]
Now continuing from Line~\ref{ln:app_v},
\begin{align}
\E[(x_0 - x_t) \mid x_t]
&=
\sum_{i < t} \E[\eta_i \mid x_t] \\
&= 
(t/\dt) \E[\eta_{t-\dt} \mid x_t] \\
&= 
(t/\dt) \E[(x_{t} - x_{t-\dt}) \mid x_t]
\end{align}
as desired.
\end{proof}
\subsection{Variance-Reduced Algorithms}
\label{app:var-red}

Here we give the ``varianced-reduced'' versions of the DDPM
training and sampling algorithms, where we train a network $g_\theta$
to approximate
\[
g_\theta(x, t) \approx \E[x_0 \mid x_t]
\]
instead of a network $f_\theta$ to approximate
\[
f_\theta(x, t) \approx \E[x_{t-\dt} \mid x_t].
\]
Via Claim~\ref{claim:var_red}, these two functions are equivalent via the transform:
\[
f_\theta(x, t) = (\dt/t) g_\theta(x, t) + (1-\dt/t)x.
\]
Plugging this relation into Pseudocode~\ref{alg:ddpm_gen} yields
the variance-reduced DDPM sampler of Pseudocode~\ref{alg:ddpm_gen_varred}.

\begin{fullwidth}

\begin{minipage}[t]{0.47\linewidth}
\begin{algorithm}[H]
\DontPrintSemicolon
\caption{DDPM train loss ($x_0$-prediction)} \label{alg:ddpm_train_varred}
\KwIn{Neural network $g_\theta$;
Sample-access to train distribution $p$.}
\KwData{Terminal variance $\sigma_q$}
\KwOut{Stochastic loss $L$}
$x_0 \gets \textrm{Sample}(p)$ \;
$t \gets \textrm{Unif}[0, 1]$ \;
$x_{t} \gets x_0 + \cN(0, \sigma_q^2 t) $ \;
$L \gets \left\Vert g_\theta(x_{t}, t) -  x_{0}\right\Vert_2^2$ \;
\Return $L$
\end{algorithm}
\end{minipage}
\hfill
\begin{minipage}[t]{0.47\linewidth}
\begin{algorithm}[H]
\DontPrintSemicolon
\caption{DDPM sampling ($x_0$-prediction)}\label{alg:ddpm_gen_varred}
\KwIn{Trained model $f_\theta$.}
\KwData{Terminal variance $\sigma_q$; step-size $\dt$.}
\KwOut{$x_0$}
$x_1 \gets \cN(0, \sigma_q^2)$ \;
\For{$t=1,~(1-\dt),(1-2\dt), \dots, \dt$}{
$\hat{\eta_t} \gets g_\theta(x_t, t) - x_t$ \;
$x_{t-\dt} \gets x_t + (1/t)\hat{\eta_t}\dt + \cN(0, \sigma_q^2\dt)$ \;
}
\Return $x_0$
\end{algorithm}
\end{minipage}

\end{fullwidth}
\subsection{Equivalence of and $x_0$- and $\eps$-prediction}
\label{append:x0_vs_eps_param}

We will discuss this in our usual simplified setup:
\begin{align*}
    x_t &= x_0 + \sigma_t \epsilon_t, \quad \sigma_t = \sigma_q \sqrt{t}, \quad \epsilon_t \sim \cN(0, 1);
\end{align*}
the scaling factors are more complex in the general case (see \citet{luo2022understanding} for VP diffusion, for example) but the idea is the same. The DDPM training algorithm \ref{alg:ddpm_train} has objective and optimal value
\begin{align*}
    \underset{\theta}{\min} &\quad \|f_\theta(x_t, t) - x_{t-\Delta t}\|_2^2, \quad f^\star_\theta(x_t, t) = \E[x_{t-\Delta t} | x_t]
\end{align*}
That is, the network $f_\theta$ to learn to predict $\E[x_{t-\Delta t} | x_t]$. However, we could instead require the network to predict other related quantities, as follows. Noting that
\begin{align*}
    &\E[x_{t-\Delta t} - x_t | x_t] 
    \overset{\text{eq. } \ref{eq:var_reduc_dt}}{=} \frac{\Delta t}{t} \E[x_0 - x_t | x_t] \equiv \frac{\Delta t}{t \sigma_t} \E[\epsilon_t | x_t] \\
    \implies & \|E[x_{t-\Delta t} - x_t | x_t] - x_{t-\Delta t}\|_2^2 = \|\frac{\Delta t}{t} (\E[x_0 | x_t] - x_0) \|_2^2 = \|\frac{\Delta t}{t \sigma_t} (\E[\epsilon_t | x_t] - \epsilon_t) \|_2^2
\end{align*}
we get the following equivalent problems: 
\begin{align*}
    \underset{\theta}{\min} &\quad \|f_\theta(x_t, t) - x_0 \|_2^2 \implies f^\star_\theta(x_t, t) = \E[x_0 | x_t], \quad \text{time-weighting } = \frac{1}{t} \\
    \underset{\theta}{\min} &\quad \|\frac{\Delta t}{t \sigma_t} (f_\theta(x_t, t) - \epsilon_t) \|_2^2 \implies f^\star_\theta(x_t, t) = \E[\epsilon_t | x_t]  \quad \text{time-weighting } = \frac{1}{t \sigma_t}.
\end{align*}

\clearpage
\newpage
\begin{fullwidth}
\bibliographystyle{plainnat}
\bibliography{refs}

\begin{thebibliography}{59}
\providecommand{\natexlab}[1]{#1}
\providecommand{\url}[1]{\texttt{#1}}
\expandafter\ifx\csname urlstyle\endcsname\relax
  \providecommand{\doi}[1]{doi: #1}\else
  \providecommand{\doi}{doi: \begingroup \urlstyle{rm}\Url}\fi

\bibitem[Advani et~al.(2020)Advani, Saxe, and Sompolinsky]{advani2020high}
Madhu~S Advani, Andrew~M Saxe, and Haim Sompolinsky.
\newblock High-dimensional dynamics of generalization error in neural networks.
\newblock \emph{Neural Networks}, 132:\penalty0 428--446, 2020.

\bibitem[Albergo et~al.(2023)Albergo, Boffi, and
  Vanden-Eijnden]{albergo2023stochastic}
Michael~S. Albergo, Nicholas~M. Boffi, and Eric Vanden-Eijnden.
\newblock Stochastic interpolants: A unifying framework for flows and
  diffusions, 2023.

\bibitem[Albergo and Vanden-Eijnden(2022)]{albergo2022building}
Michael~Samuel Albergo and Eric Vanden-Eijnden.
\newblock Building normalizing flows with stochastic interpolants.
\newblock In \emph{The Eleventh International Conference on Learning
  Representations}, 2022.

\bibitem[Anderson(1982)]{anderson1982reverse}
Brian~DO Anderson.
\newblock Reverse-time diffusion equation models.
\newblock \emph{Stochastic Processes and their Applications}, 12\penalty0
  (3):\penalty0 313--326, 1982.

\bibitem[Carlini et~al.(2023)Carlini, Hayes, Nasr, Jagielski, Sehwag, Tramer,
  Balle, Ippolito, and Wallace]{carlini2023extracting}
Nicolas Carlini, Jamie Hayes, Milad Nasr, Matthew Jagielski, Vikash Sehwag,
  Florian Tramer, Borja Balle, Daphne Ippolito, and Eric Wallace.
\newblock Extracting training data from diffusion models.
\newblock In \emph{32nd USENIX Security Symposium (USENIX Security 23)}, pages
  5253--5270, 2023.

\bibitem[Chan(2024)]{chan2024tutorial}
Stanley~H. Chan.
\newblock Tutorial on diffusion models for imaging and vision, 2024.

\bibitem[Chen et~al.(2023)Chen, Lee, and Lu]{chen2023improved}
Hongrui Chen, Holden Lee, and Jianfeng Lu.
\newblock Improved analysis of score-based generative modeling: User-friendly
  bounds under minimal smoothness assumptions.
\newblock In \emph{International Conference on Machine Learning}, pages
  4735--4763. PMLR, 2023.

\bibitem[Chen et~al.(2022)Chen, Chewi, Li, Li, Salim, and
  Zhang]{chen2022sampling}
Sitan Chen, Sinho Chewi, Jerry Li, Yuanzhi Li, Adil Salim, and Anru Zhang.
\newblock Sampling is as easy as learning the score: theory for diffusion
  models with minimal data assumptions.
\newblock In \emph{The Eleventh International Conference on Learning
  Representations}, 2022.

\bibitem[Chen et~al.(2024{\natexlab{a}})Chen, Chewi, Lee, Li, Lu, and
  Salim]{chen2024probability}
Sitan Chen, Sinho Chewi, Holden Lee, Yuanzhi Li, Jianfeng Lu, and Adil Salim.
\newblock The probability flow ode is provably fast.
\newblock \emph{Advances in Neural Information Processing Systems}, 36,
  2024{\natexlab{a}}.

\bibitem[Chen et~al.(2024{\natexlab{b}})Chen, Kontonis, and
  Shah]{chen2024learning}
Sitan Chen, Vasilis Kontonis, and Kulin Shah.
\newblock Learning general gaussian mixtures with efficient score matching.
\newblock \emph{arXiv preprint arXiv:2404.18893}, 2024{\natexlab{b}}.

\bibitem[Das(2024)]{das2024buildingdiffusionmodels}
Ayan Das.
\newblock Building diffusion model's theory from ground up.
\newblock In \emph{ICLR Blogposts 2024}, 2024.
\newblock URL
  \url{https://iclr-blogposts.github.io/2024/blog/diffusion-theory-from-scratch/}.
\newblock
  https://iclr-blogposts.github.io/2024/blog/diffusion-theory-from-scratch/.

\bibitem[De~Bortoli(2022)]{de2022convergence}
Valentin De~Bortoli.
\newblock Convergence of denoising diffusion models under the manifold
  hypothesis.
\newblock \emph{arXiv preprint arXiv:2208.05314}, 2022.

\bibitem[De~Bortoli et~al.(2021)De~Bortoli, Thornton, Heng, and
  Doucet]{de2021diffusion}
Valentin De~Bortoli, James Thornton, Jeremy Heng, and Arnaud Doucet.
\newblock Diffusion schr{\"o}dinger bridge with applications to score-based
  generative modeling.
\newblock \emph{Advances in Neural Information Processing Systems},
  34:\penalty0 17695--17709, 2021.

\bibitem[Dieleman(2023)]{dieleman2023perspectives}
Sander Dieleman.
\newblock Perspectives on diffusion, 2023.
\newblock URL \url{https://sander.ai/2023/07/20/perspectives.html}.

\bibitem[Duan(2023)]{tonydiff}
Tony Duan.
\newblock Diffusion models from scratch, 2023.
\newblock URL \url{https://www.tonyduan.com/diffusion/index.html}.

\bibitem[Eldan(2024)]{eldan_blog_sde}
Ronen Eldan.
\newblock Lecture notes - from stochastic calculus to geometric inequalities,
  2024.
\newblock URL \url{https://www.wisdom.weizmann.ac.il/~ronene/GFANotes.pdf}.

\bibitem[Esser et~al.(2024)Esser, Kulal, Blattmann, Entezari, M{\"u}ller,
  Saini, Levi, Lorenz, Sauer, Boesel, et~al.]{esser2024scaling}
Patrick Esser, Sumith Kulal, Andreas Blattmann, Rahim Entezari, Jonas
  M{\"u}ller, Harry Saini, Yam Levi, Dominik Lorenz, Axel Sauer, Frederic
  Boesel, et~al.
\newblock Scaling rectified flow transformers for high-resolution image
  synthesis.
\newblock \emph{arXiv preprint arXiv:2403.03206}, 2024.

\bibitem[Evans(2012)]{evans2012introduction}
Lawrence~C Evans.
\newblock \emph{An introduction to stochastic differential equations},
  volume~82.
\newblock American Mathematical Soc., 2012.

\bibitem[Fjelde et~al.(2024)Fjelde, Mathieu, and Dutordoir]{mathieu2024flow}
Tor Fjelde, Emile Mathieu, and Vincent Dutordoir.
\newblock An introduction to flow matching, January 2024.
\newblock URL
  \url{https://mlg.eng.cam.ac.uk/blog/2024/01/20/flow-matching.html}.

\bibitem[Gu et~al.(2023)Gu, Du, Pang, Li, Lin, and Wang]{gu2023memorization}
Xiangming Gu, Chao Du, Tianyu Pang, Chongxuan Li, Min Lin, and Ye~Wang.
\newblock On memorization in diffusion models.
\newblock \emph{arXiv preprint arXiv:2310.02664}, 2023.

\bibitem[Ho et~al.(2020)Ho, Jain, and Abbeel]{ho2020denoising}
Jonathan Ho, Ajay Jain, and Pieter Abbeel.
\newblock Denoising diffusion probabilistic models.
\newblock \emph{Advances in neural information processing systems},
  33:\penalty0 6840--6851, 2020.

\bibitem[Kadkhodaie et~al.(2024)Kadkhodaie, Guth, Simoncelli, and
  Mallat]{kadkhodaie2024generalization}
Zahra Kadkhodaie, Florentin Guth, Eero~P Simoncelli, and St{\'e}phane Mallat.
\newblock Generalization in diffusion models arises from geometry-adaptive
  harmonic representations.
\newblock In \emph{The Twelfth International Conference on Learning
  Representations}, 2024.
\newblock URL \url{https://openreview.net/forum?id=ANvmVS2Yr0}.

\bibitem[Karras et~al.(2022)Karras, Aittala, Aila, and
  Laine]{karras2022elucidating}
Tero Karras, Miika Aittala, Timo Aila, and Samuli Laine.
\newblock Elucidating the design space of diffusion-based generative models,
  2022.

\bibitem[Kingma and Gao(2023)]{kingma2023understanding}
Diederik~P Kingma and Ruiqi Gao.
\newblock Understanding diffusion objectives as the {ELBO} with simple data
  augmentation.
\newblock In \emph{Thirty-seventh Conference on Neural Information Processing
  Systems}, 2023.
\newblock URL \url{https://openreview.net/forum?id=NnMEadcdyD}.

\bibitem[Kloeden and Platen(2011)]{kloeden2011numerical}
P.E. Kloeden and E.~Platen.
\newblock \emph{Numerical Solution of Stochastic Differential Equations}.
\newblock Stochastic Modelling and Applied Probability. Springer Berlin
  Heidelberg, 2011.
\newblock ISBN 9783540540625.
\newblock URL \url{https://books.google.com/books?id=BCvtssom1CMC}.

\bibitem[Lee et~al.(2023)Lee, Lu, and Tan]{lee2023convergence}
Holden Lee, Jianfeng Lu, and Yixin Tan.
\newblock Convergence of score-based generative modeling for general data
  distributions.
\newblock In \emph{International Conference on Algorithmic Learning Theory},
  pages 946--985. PMLR, 2023.

\bibitem[Lin et~al.(2024)Lin, Wang, and Yang]{lin2024sdxl}
Shanchuan Lin, Anran Wang, and Xiao Yang.
\newblock Sdxl-lightning: Progressive adversarial diffusion distillation.
\newblock \emph{arXiv preprint arXiv:2402.13929}, 2024.

\bibitem[Lipman et~al.(2023)Lipman, Chen, Ben-Hamu, Nickel, and
  Le]{lipman2023flow}
Yaron Lipman, Ricky T.~Q. Chen, Heli Ben-Hamu, Maximilian Nickel, and Matthew
  Le.
\newblock Flow matching for generative modeling.
\newblock In \emph{The Eleventh International Conference on Learning
  Representations}, 2023.
\newblock URL \url{https://openreview.net/forum?id=PqvMRDCJT9t}.

\bibitem[Liu et~al.(2022{\natexlab{a}})Liu, Gong, et~al.]{liu2022flow}
Xingchao Liu, Chengyue Gong, et~al.
\newblock Flow straight and fast: Learning to generate and transfer data with
  rectified flow.
\newblock In \emph{The Eleventh International Conference on Learning
  Representations}, 2022{\natexlab{a}}.

\bibitem[Liu et~al.(2022{\natexlab{b}})Liu, Wu, Ye, and Liu]{liu2022let}
Xingchao Liu, Lemeng Wu, Mao Ye, and Qiang Liu.
\newblock Let us build bridges: Understanding and extending diffusion
  generative models, 2022{\natexlab{b}}.

\bibitem[Lu et~al.(2022{\natexlab{a}})Lu, Zhou, Bao, Chen, Li, and
  Zhu]{lu2022dpm}
Cheng Lu, Yuhao Zhou, Fan Bao, Jianfei Chen, Chongxuan Li, and Jun Zhu.
\newblock Dpm-solver: A fast ode solver for diffusion probabilistic model
  sampling in around 10 steps.
\newblock \emph{Advances in Neural Information Processing Systems},
  35:\penalty0 5775--5787, 2022{\natexlab{a}}.

\bibitem[Lu et~al.(2022{\natexlab{b}})Lu, Zhou, Bao, Chen, Li, and
  Zhu]{lu2022dpmplusplus}
Cheng Lu, Yuhao Zhou, Fan Bao, Jianfei Chen, Chongxuan Li, and Jun Zhu.
\newblock Dpm-solver++: Fast solver for guided sampling of diffusion
  probabilistic models.
\newblock \emph{arXiv preprint arXiv:2211.01095}, 2022{\natexlab{b}}.

\bibitem[Luo(2022)]{luo2022understanding}
Calvin Luo.
\newblock Understanding diffusion models: A unified perspective, 2022.

\bibitem[McAllester(2023)]{mcallester2023mathematics}
David McAllester.
\newblock On the mathematics of diffusion models, 2023.

\bibitem[Montanari(2023)]{montanari2023sampling}
Andrea Montanari.
\newblock Sampling, diffusions, and stochastic localization, 2023.

\bibitem[Peluchetti(2022)]{peluchetti2022nondenoising}
Stefano Peluchetti.
\newblock Non-denoising forward-time diffusions, 2022.
\newblock URL \url{https://openreview.net/forum?id=oVfIKuhqfC}.

\bibitem[Permenter and Yuan(2023)]{permenter2023interpreting}
Frank Permenter and Chenyang Yuan.
\newblock Interpreting and improving diffusion models using the euclidean
  distance function.
\newblock \emph{arXiv preprint arXiv:2306.04848}, 2023.

\bibitem[Peyr{\'e}(2023)]{peyre2023denoising}
Gabriel Peyr{\'e}.
\newblock Denoising diffusion models, 2023.
\newblock URL
  \url{https://mathematical-tours.github.io/book-sources/optim-ml/OptimML-DiffusionModels.pdf}.

\bibitem[Pooladian et~al.(2023)Pooladian, Ben-Hamu, Domingo-Enrich, Amos,
  Lipman, and Chen]{pooladian2023multisample}
Aram-Alexandre Pooladian, Heli Ben-Hamu, Carles Domingo-Enrich, Brandon Amos,
  Yaron Lipman, and Ricky~TQ Chen.
\newblock Multisample flow matching: Straightening flows with minibatch
  couplings.
\newblock In \emph{International Conference on Machine Learning}, pages
  28100--28127. PMLR, 2023.

\bibitem[Ribeiro and Glocker(2024)]{ribeiro2024demystifying}
Fabio De~Sousa Ribeiro and Ben Glocker.
\newblock Demystifying variational diffusion models, 2024.

\bibitem[Salimans and Ho(2022)]{salimans2022progressive}
Tim Salimans and Jonathan Ho.
\newblock Progressive distillation for fast sampling of diffusion models.
\newblock \emph{arXiv preprint arXiv:2202.00512}, 2022.

\bibitem[Sauer et~al.(2024)Sauer, Boesel, Dockhorn, Blattmann, Esser, and
  Rombach]{sauer2024fast}
Axel Sauer, Frederic Boesel, Tim Dockhorn, Andreas Blattmann, Patrick Esser,
  and Robin Rombach.
\newblock Fast high-resolution image synthesis with latent adversarial
  diffusion distillation.
\newblock \emph{arXiv preprint arXiv:2403.12015}, 2024.

\bibitem[Sohl{-}Dickstein et~al.(2015)Sohl{-}Dickstein, Weiss, Maheswaranathan,
  and Ganguli]{OGdiffusion}
Jascha Sohl{-}Dickstein, Eric~A. Weiss, Niru Maheswaranathan, and Surya
  Ganguli.
\newblock Deep unsupervised learning using nonequilibrium thermodynamics.
\newblock \emph{CoRR}, abs/1503.03585, 2015.
\newblock URL \url{http://arxiv.org/abs/1503.03585}.

\bibitem[Somepalli et~al.(2023)Somepalli, Singla, Goldblum, Geiping, and
  Goldstein]{somepalli2023diffusion}
Gowthami Somepalli, Vasu Singla, Micah Goldblum, Jonas Geiping, and Tom
  Goldstein.
\newblock Diffusion art or digital forgery? investigating data replication in
  diffusion models.
\newblock In \emph{Proceedings of the IEEE/CVF Conference on Computer Vision
  and Pattern Recognition}, pages 6048--6058, 2023.

\bibitem[Song et~al.(2021)Song, Meng, and Ermon]{song2021denoising}
Jiaming Song, Chenlin Meng, and Stefano Ermon.
\newblock Denoising diffusion implicit models.
\newblock In \emph{International Conference on Learning Representations}, 2021.
\newblock URL \url{https://openreview.net/forum?id=St1giarCHLP}.

\bibitem[Song et~al.(2023{\natexlab{a}})Song, Meng, and Vahdat]{cvpr2023}
Jiaming Song, Chenlin Meng, and Arash Vahdat.
\newblock Cvpr 2023 tutorial: Denoising diffusion models: A generative learning
  big bang, 2023{\natexlab{a}}.
\newblock URL \url{https://cvpr2023-tutorial-diffusion-models.github.io}.

\bibitem[Song(2021)]{songblog2021generative}
Yang Song.
\newblock Generative modeling by estimating gradients of the data distribution,
  2021.
\newblock URL \url{https://yang-song.net/blog/2021/score/}.

\bibitem[Song(2023)]{songVideo}
Yang Song.
\newblock Diffusion and score-based generative models, 2023.
\newblock URL \url{https://www.youtube.com/watch?v=wMmqCMwuM2Q}.

\bibitem[Song and Ermon(2019)]{song2019generative}
Yang Song and Stefano Ermon.
\newblock Generative modeling by estimating gradients of the data distribution.
\newblock \emph{Advances in neural information processing systems}, 32, 2019.

\bibitem[Song et~al.(2020)Song, Sohl-Dickstein, Kingma, Kumar, Ermon, and
  Poole]{song2020score}
Yang Song, Jascha Sohl-Dickstein, Diederik~P Kingma, Abhishek Kumar, Stefano
  Ermon, and Ben Poole.
\newblock Score-based generative modeling through stochastic differential
  equations.
\newblock \emph{arXiv preprint arXiv:2011.13456}, 2020.
\newblock URL \url{https://arxiv.org/pdf/2011.13456.pdf}.

\bibitem[Song et~al.(2023{\natexlab{b}})Song, Dhariwal, Chen, and
  Sutskever]{song2023consistency}
Yang Song, Prafulla Dhariwal, Mark Chen, and Ilya Sutskever.
\newblock Consistency models.
\newblock \emph{arXiv preprint arXiv:2303.01469}, 2023{\natexlab{b}}.

\bibitem[Stark et~al.(2024)Stark, Jing, Wang, Corso, Berger, Barzilay, and
  Jaakkola]{stark2024dirichlet}
Hannes Stark, Bowen Jing, Chenyu Wang, Gabriele Corso, Bonnie Berger, Regina
  Barzilay, and Tommi Jaakkola.
\newblock Dirichlet flow matching with applications to dna sequence design,
  2024.

\bibitem[Tong et~al.(2023)Tong, Malkin, Fatras, Atanackovic, Zhang, Huguet,
  Wolf, and Bengio]{tong2023simulation}
Alexander Tong, Nikolay Malkin, Kilian Fatras, Lazar Atanackovic, Yanlei Zhang,
  Guillaume Huguet, Guy Wolf, and Yoshua Bengio.
\newblock Simulation-free schr$\backslash$" odinger bridges via score and flow
  matching.
\newblock \emph{arXiv preprint arXiv:2307.03672}, 2023.

\bibitem[Turner(2021)]{turner_blog}
Angus Turner.
\newblock Diffusion models as a kind of vae, June 2021.
\newblock URL
  \url{https://angusturner.github.io/generative_models/2021/06/29/diffusion-probabilistic-models-I.html}.

\bibitem[Winkler(2021)]{winkler_blog_reverse}
Ludwig Winkler.
\newblock Reverse time stochastic differential equations [for generative
  modeling], 2021.
\newblock URL \url{https://ludwigwinkler.github.io/blog/ReverseTimeAnderson/}.

\bibitem[Winkler(2023)]{winkler_blog_fp}
Ludwig Winkler.
\newblock Fokker, planck, and ito, 2023.
\newblock URL \url{https://ludwigwinkler.github.io/blog/FokkerPlanck/}.

\bibitem[Xu et~al.(2023)Xu, Zhao, Xiao, and Hou]{xu2023ufogen}
Yanwu Xu, Yang Zhao, Zhisheng Xiao, and Tingbo Hou.
\newblock Ufogen: You forward once large scale text-to-image generation via
  diffusion gans.
\newblock \emph{arXiv preprint arXiv:2311.09257}, 2023.

\bibitem[Yuan(2024)]{CYblog}
Chenyang Yuan.
\newblock Diffusion models from scratch, from a new theoretical perspective,
  2024.
\newblock URL \url{https://www.chenyang.co/diffusion.html}.

\bibitem[Zhang and Chen(2023)]{zhang2023fast}
Qinsheng Zhang and Yongxin Chen.
\newblock Fast sampling of diffusion models with exponential integrator.
\newblock In \emph{The Eleventh International Conference on Learning
  Representations}, 2023.
\newblock URL \url{https://openreview.net/forum?id=Loek7hfb46P}.

\end{thebibliography}
\end{fullwidth}

\end{document}